\pdfoutput=1
\PassOptionsToPackage{dvipsnames}{xcolor}
\documentclass[11pt]{article}

\usepackage[]{ACL2023}

\usepackage{times}
\usepackage{latexsym}

\usepackage[T1]{fontenc}

\usepackage[utf8]{inputenc}

\usepackage{microtype}

\usepackage{inconsolata}

\usepackage{CJKutf8}
\usepackage{tikz}
\usepackage{calc}
\usetikzlibrary{patterns,shapes,snakes,calendar,calc,backgrounds,folding,arrows,positioning,math}
\usepackage{mathrsfs}
\usepackage{paralist}
\usepackage{booktabs}
\usepackage{multirow}
\usepackage{pgfplots}
\usepackage{scalefnt}
\usepackage{microtype}
\usepackage{amsthm,amsmath,amsfonts,amssymb}
\usepackage[lined,boxed,commentsnumbered,ruled]{algorithm2e}
\usepackage{bm}
\usepackage{colortbl}
\usepackage{array}
\usepackage{stfloats}
\usepackage{siunitx}
\usepackage[ruled]{algorithm2e}
\usepackage{subfigure}
\usepackage{pgfplots}
\usepackage[shortcuts]{extdash}
\usepackage{makecell}
\usepackage{caption}

\usepackage{arydshln}

\usepackage{paralist}

\usepackage{tcolorbox}

\usepackage{adjustbox}

\usepgfplotslibrary{groupplots}
\title{Generating Visual Spatial Description via Holistic 3D Scene Understanding}

\author{Yu Zhao$^{1,2}$, \quad  Hao Fei$^2$\Thanks{ Corresponding author: Hao Fei}, \quad  Wei Ji$^2$, \quad Jianguo Wei$^1$\\
\bf Meishan Zhang$^3$, \quad  Min Zhang$^3$,  \quad Tat-Seng Chua$^2$ \\
  $^1$ College of Intelligence and Computing, Tianjin University, China \\
  $^2$ Sea-NExT Joint Lab, National University of Singapore, Singapore\\
  $^3$ Harbin Institute of Technology (Shenzhen), China \\
  \texttt{\{zhaoyucs,jianguowei\}@tju.edu.cn}, \,  \texttt{\{haofei37,jiwei,dcscts\}@nus.edu.sg}\\
   \texttt{mason.zms@gmail.com}, \,  \texttt{zhangmin2021@hit.edu.cn} \\}

\begin{document}
\maketitle
\begin{abstract}
Visual spatial description (VSD) aims to generate texts that describe the spatial relations of the given objects within images.
Existing VSD work merely models the 2D geometrical vision features, thus inevitably falling prey to the problem of skewed spatial understanding of target objects.
In this work, we investigate the incorporation of 3D scene features for VSD.
With an external 3D scene extractor, we obtain the 3D objects and scene features for input images, based on which we construct a target object-centered 3D spatial scene graph (\textsc{Go3D}-S$^2$G), such that we model the spatial semantics of target objects within the holistic 3D scenes.
Besides, we propose a scene subgraph selecting mechanism, sampling topologically-diverse subgraphs from \textsc{Go3D}-S$^2$G, where the diverse local structure features are navigated to yield spatially-diversified text generation.
Experimental results on two VSD datasets demonstrate that our framework outperforms the baselines significantly, especially improving on the cases with complex visual spatial relations.
Meanwhile, our method can produce more spatially-diversified generation.
Code is available at \url{https://github.com/zhaoyucs/VSD}.
\end{abstract}
\begin{CJK*}{UTF8}{gbsn}

\section{Introduction}

\vspace{-1.5pt}
Visual spatial description is a newly emerged vision-language task, which aims to generate a textual descriptive sentence of the spatial relationship between two target visual objects in a given image \cite{zhaoyu-vsd}.
VSD falls into the category of image-to-text generation, while in particular focusing on the visual spatial semantics understanding, which has great values on the real-world human-computer interaction (HCI) applications \cite{HeuserAS20}, e.g., automatic navigation \cite{PendaoM21,wang2023sar}, personal assistance \cite{VanhooydonckDHPVBN10}, and unmanned manipulation \cite{CastamanTABGCMB21,wang2023entropy,wang2022recognition}.

\begin{figure}[!t]
    \centering
    \includegraphics[width=0.98\columnwidth]{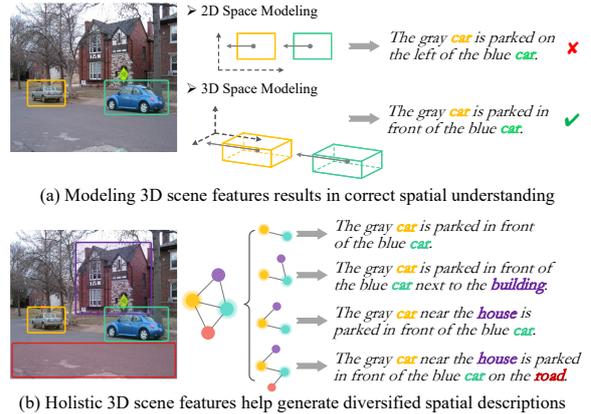}
    \vspace{-1mm}
    \caption{Examples of the visual spatial description.}
    \label{fig:intro}
    \vspace{-5mm}
\end{figure}

\vspace{-1.5pt}
\citet{zhaoyu-vsd} pioneer the VSD task by manually annotating the spatial descriptions to the images based on the visual spatial classification datasets \cite{KrishnaZGJHKCKL17}.
Also they solve VSD as a general image-to-text (I2T) task via vision-language pre-trained models (VL-PTMs), i.e., inputting images and outputting texts.
However, modeling VSD as a regular I2T job with open-ended VL-PTMs can be problematic.
Unlike the existing I2T tasks, such as image captioning \cite{DBLP:conf/cvpr/VinyalsTBE15}, verb-specific semantic roles (VSR) guided captioning \cite{0016J0021} and visual question answering (VQA) \cite{DBLP:conf/iccv/AntolALMBZP15} that focus on the \emph{content semantics} understanding, VSD emphasizes more on the \emph{spatial semantics} reasoning, according to its definition.
Thus, directly adapting VSD with general-purpose VL-PTMs will lead to inferior task performances.
We note that there are at least two observations that should be taken into account for VSD enhancement.

\begin{figure*}[!t]
    \centering
    \includegraphics[width=0.98\linewidth]{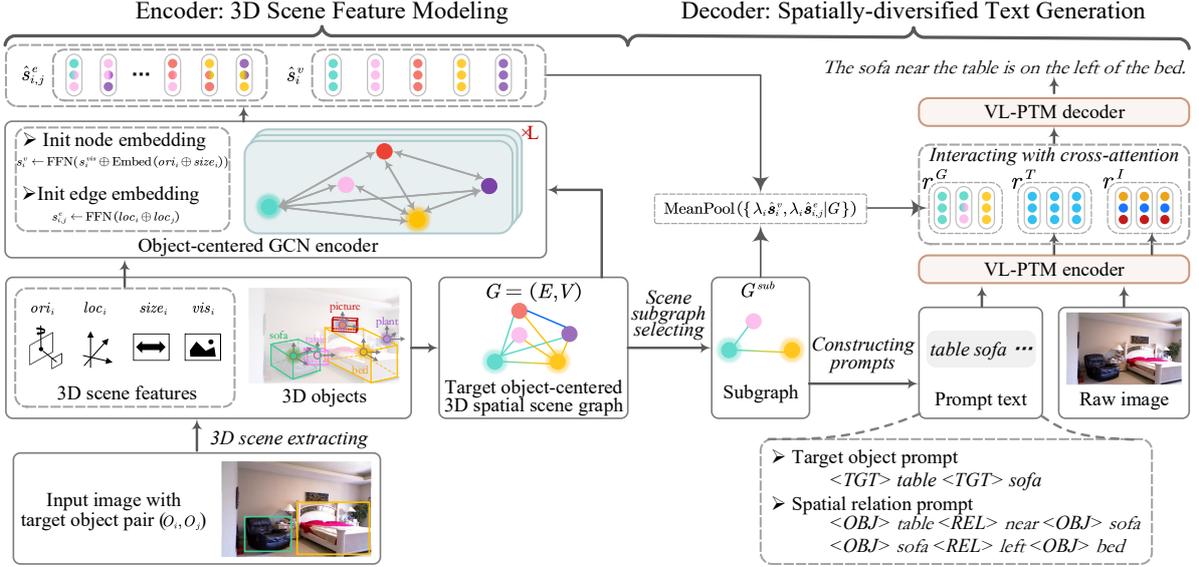}
    \vspace{-1mm}
    \caption{
    The overview of our proposed framework.
    }
    \label{fig:2}
    \vspace{-4mm}
\end{figure*}

\vspace{-1.5pt}
From the image encoding aspect, it is critical to model the holistic 3D scene semantics of the input image.
In \citet{zhaoyu-vsd}, their VL-PTM-based methods model the visual geometrical features at merely 2D flat space (e.g., superficial features).
Yet directly perceiving objects from the first-person perspective will inevitably result in skewed angle of view and biased spatial understanding, and thus fail to handle complex cases (e.g. layout overlap, perspective illusion) or generate incorrect descriptions.
For example as in Figure \ref{fig:intro}(a), with the 2D-level visual understanding, the spatial relation between two cars is wrongly decided.
As a reference, we human always first project the visual contents into the 3D space and reckon the scene layout and object attributes (e.g., depth, shapes, camera poses), and then narrate the spatial relations based on such holistic 3D clues.

\vspace{-1.5pt}
From the text decoding aspect, it is necessary yet challenging to generate diverse sentences of the object pair relation.
In generic I2T task, prior methods strengthen the text diversification by equipping with beam search \cite{VijayakumarCSSL18} or integrating external knowledge \cite{yu-etal-2022-diversifying}.
Different from the generic diversified generation, VSD requires the diversification with respect to the spatial descriptions, rather than the diverse linguistics.
We can again place the emphasis on the modeling of holistic 3D scene features.
For example, with a precise understanding of the spatial relations, it is both viable to generate `A is \emph{on the left of} B' or `B is \emph{on the right of} A'.
Also, as illustrated in Figure \ref{fig:intro}(b), by comprehensively modeling the surrounding relations of the neighbor objects connecting to the target objects in the holistic scene, more spatially-diverse texts can be yielded via different path traversing.

\vspace{-1.5pt}
In this paper, we propose enhancing VSD by modeling the holistic 3D scene semantics.
We build an encoder-decoder VSD framework (cf. Figure \ref{fig:2}), where the encoder learns the 3D spatial features, and the decoder generates spatially-diversified descriptions based on the spatial semantic features.
Specifically, at encoding side, we first employ an off-the-shelf 3D scene extractor \cite{DBLP:conf/cvpr/NieHGZCZ20} to produce 3D objects and the corresponding scene features (i.e., layout, location, size and visual features) for the input monocular RGB image, via which we build a tar\underline{g}et \underline{o}bject-centered \underline{3D} \underline{s}patial \underline{s}cene \underline{g}raph (namely, \textsc{Go3D}-S$^2$G).
We then present an object-centered graph convolutional network (\textsc{OcGCN}) to encode the \textsc{Go3D}-S$^2$G.
At decoding side, we devise a scene subgraph selecting (S$^3$) mechanism to sample topologically-diverse object-neighboring subgraphs from \textsc{Go3D}-S$^2$G, which allows to generate descriptions focusing on the near surroundings of target object.
Based on the sampled subgraphs, we then create prompt texts to ground the focused objects and their prototype directions.
Finally, a backbone VL-PTM is used to encode the prompt texts, input images as well as 3D scene features, then to produce VSD texts.

\vspace{-0.5pt}
We experiment on two versions of VSD datasets \cite{zhaoyu-vsd}, where one is with simple annotations and one has more complex and human-friendly descriptions.
The results indicate that our system outperforms the best baseline with significant margins, where our method especially improves on the complex cases, such as layout-overlapped and irregularly-posed objects.
We further reveal how the 3D scene graph modeling as well as the S$^3$ mechanism facilitate the task, and also quantify the influence of the external 3D scene extractor.
All in all, this work contributes by verifying that modeling the 3D scene of 2D image
helps the understanding of visual spatial semantics.

\vspace{-2mm}
\section{Methodology}

\vspace{-1mm}
\paragraph{Problem Definition}
Given an image $I$ with two object proposals <$O_1, O_2$> in $I$, VSD generates a sequence of words $S=\{w_1,...,w_n\}$ that describes the spatial relationship between $O_1$ and $O_2$. 
The input $O_1$ and $O_2$ contain the object tags and their 2D location coordinates. 
Different from image captioning, the generated sentences of VSD must directly or indirectly express the spatial relation between the target objects.

\vspace{-1mm}
\paragraph{Overall Framework} \label{section:overall}

As shown in Figure \ref{fig:2}, our framework (namely \textsc{3Dvsd}) is built upon an encoder-decoder paradigm, where the encoder is responsible for the 3D scene feature modeling, and the decoder generates spatially-diversified descriptions based on the spatial semantic features learned from encoder.

\vspace{-2mm}
\subsection{Encoder: 3D Scene Feature Modeling}

We first extract 3D scene features via an external extractor.
Then we build a target object-centered 3D spatial scene graph (\textsc{Go3D}-S$^2$G), which is encoded and propagated with an object-centered GCN (\textsc{OcGCN}).

\begin{table}[t]
\fontsize{10}{12.5}\selectfont
\setlength{\tabcolsep}{1.mm}
\renewcommand\arraystretch{0.6}
\begin{center}
\resizebox{0.98\columnwidth}{!}{
\begin{tabular}{ll}
\hline
\multirow{2}{*}{$vis_i$} & \multirow{2}{*}{The flatted ROI feature of object $i$.}\\ \\
\multirow{2}{*}{$size_i$} & \multirow{2}{*}{The length, width, height of object $i$.}\\ \\
\multirow{2}{*}{$loc_i$} & \multirow{2}{*}{The relative centroid coordinates of object $i$.} \\ \\
\multirow{2}{*}{$ori_i$} & The rotation value of three degrees of freedom \\
& of object $i$.\\
\hline
\end{tabular}
}
\vspace{-1mm}
\caption{Summary of the 3D scene features.
}
\label{tab:3D-features}
\end{center}
\vspace{-6mm}
\end{table}

\vspace{-1mm}
\paragraph{Extracting 3D Scene Features}

We adopt the 3D scene extractor as in \citet{DBLP:conf/cvpr/NieHGZCZ20}, which is a joint layout estimator and 3D object detector. 
It first processes the 2D object detection for the input RGB image, based on which the 3D relative coordinates (location) and pose parameters of all the objects will be estimated.
Formally, we set up the world system located at the camera center with its vertical axis perpendicular to the floor, and its forward axis toward the camera, such that the camera pose $R(\beta, \gamma)$ can be decided by the pitch and roll angles $(\beta, \gamma)$.
In the world system, an object $O_i$ can be determined by a 3D center $loc_i \in \mathbb{R}^3$, spatial size $size_i \in \mathbb{R}^3$, orientation angle $ori_i \in [-\pi, \pi)^3$. 
Finally, we obtain the $loc_i$, $size_i$, $ori_i$ and the region-of-interest (RoI) $vis_i$ (with its representation $r_i^{vis}$) of each 3D object, which is summarized in Table \ref{tab:3D-features}.
We extend the 3D scene generating details at Appendix \ref{3D Scene Extracting}.

\vspace{-1mm}
\paragraph{Constructing Target Object-centered 3D Spatial Scene Graph}

Based on the 3D objects and the corresponding 3D scene features, we now construct the \textsc{Go3D}-S$^2$G.
The graph is centered on the two target objects, placing the focus on the spatial relationships between the target objects and their surrounding neighbor objects in the scene.
Technically, we denote \textsc{Go3D}-S$^2$G as $G$ =$(E,V)$, where 
$V$ is the set of 3D nodes $v_i$ (i.e., 3D objects).
Note that as the input images are likely to contain some noisy objects that are less-informative to the task, we remove those objects by comparing their confidence $f_i$ (i.e., the logit from the object detector) with a threshold $p$.
$E$ is the set of edges $e_{i,j}$, consisting three types:
\begin{compactitem}
    \item \textbf{Target-pair edge}, which connects two given target objects.

    \item \textbf{Target-surrounding edge}, which connects each target object to all their surrounding non-target objects.

    \item \textbf{Near-neighbor edge}, which connects those non-target objects in near neighbor that may have implicit correlations between each other.
    We build the edges by calculating their coordinates ($loc_i$), with those values larger than a pre-defined threshold $d$ as valid edges.

\end{compactitem}

\begin{figure}[!t]
    \centering
    \includegraphics[width=0.94\linewidth]{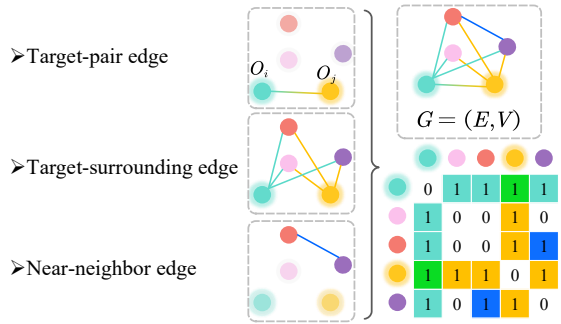}
    \vspace{-1mm}
    \caption{Three types of edges of \textsc{Go3D}-S$^2$G.}
    \label{fig:Gone3D-SSG}
\end{figure}

\noindent The edge $e_{i,j}$=1 when there is an edge between $v_i$ and $v_j$.
Figure \ref{fig:Gone3D-SSG} illustrates the edge constructions.

\vspace{-1mm}
\paragraph{Encoding Graph with Object-centered GCN}

While graph convolutional network (GCN) \cite{marcheggiani-titov-2017-encoding} has been shown effective for aggregating graph data, it may fail to model the centralization of the target objects of \textsc{Go3D}-S$^2$G structure (as GCN treats all nodes equally).
Thus, we devise an object-centered GCN, which advances in modeling both the edge features and the target objects.
\textsc{OcGCN} first creates initial representations of node $\bm{s}^v_i$ and edge $\bm{s}^e_{i,j}$.
\setlength\abovedisplayskip{3pt}
\setlength\belowdisplayskip{3pt}
\begin{equation} \small \label{equation:v}
\begin{split}
\bm{s}^{pose}_i &= \text{Embed}({ori}_i \oplus {size}_i) \,, \\
\bm{s}^v_i &= \text{FFN}(\bm{s}^{vis}_i \oplus \bm{s}^{pose}_i)\,, \\
\bm{s}^e_{i,j} &= \text{FFN}(loc_{i} \oplus loc_{j})\,, \\
\end{split}
\end{equation}
where Embed() is the looking-up operation, FFN() is the non-linear feedforward network.

Then, \textsc{OcGCN} updates the \textsc{Go3D}-S$^2$G:
\setlength\abovedisplayskip{3pt}
\setlength\belowdisplayskip{3pt}
\begin{equation}\small \label{OcGCN}
\begin{split}
\bm{s}^v_i &= \sigma(\sum_{j=1} \gamma_{i,j}(\bm{W}_a \cdot [\bm{s}^v_{i^{'}} ; \bm{s}^e_{i,j} ; \bm{s}^v_{t}] ) ) \,, \\
\gamma_{i,j}  &= \frac{e_{i,j} \cdot \exp(\bm{W}_b (\bm{s}^v_{j^{'}} \oplus e_{i,j} \oplus \bm{s}^v_{t}))}{\sum_{t=1} e_{i,t} \cdot \exp(\bm{W}_b (\bm{s}^v_{t^{'}} \oplus e_{i,t}\oplus \bm{s}^v_{t}))} \,, \\
\end{split}
\end{equation}
where $\bm{s}^v_{i^{'}}$ is the node representation of last layer, as \textsc{OcGCN} has total $L$ layers.
$\bm{s}^v_{t}$=$\bm{s}^v_{O_1}\oplus\bm{s}^v_{O_2}$ is the summary of the two target objects.
$[;]$ is the concatenation operation.
$\bm{W}_a,\bm{W}_b,b$ are learnable parameters.
The weight $\gamma_{i,j}^l$ reflects the contribution of each object when propagating the spatial attributes towards target objects.

\begin{figure}[t]
    \centering
    \includegraphics[width=0.98\linewidth]{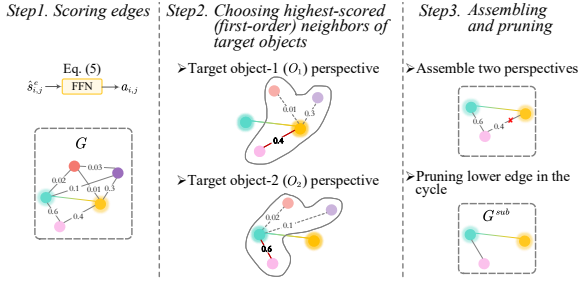}
    \vspace{-1mm}
    \caption{Scene subgraph selecting mechanism.}
    \label{fig:3}
\end{figure}

\vspace{-1mm}
\subsection{Decoder: Spatially-diversified Text Generation}

\vspace{-1mm}
In decoding stage, we use a VL-PLM to generate VSD texts, as shown in Figure \ref{fig:2}.
We first perform scene subgraph selection over \textsc{Go3D}-S$^2$G, where the diverse local structures lead to spatially-diversified text generation.
Also we create prompt texts to hint model to generate relevant contents.

\vspace{-1mm}
\paragraph{Scene Subgraph Selecting}
As cast earlier, we can make use of the neighbor non-target objects of the two target objects in the scene as type of `bridges' to diversify the generation.
Thus, we propose a scene subgraph selecting (namely, S$^3$) mechanism to sample sub-structures of \textsc{Go3D}-S$^2$G.

Concretely, S$^3$ contains three steps, as illustrated in Figure \ref{fig:3}. 
First, we calculate the connecting-strength score for each edges in \textsc{Go3D}-S$^2$G via a simple FFN transformation: $a_{i,j}$=FFN($\hat{s}^e_{i,j}$), where $\hat{s}^e_{i,j}$ is the edge representation of final-layer \textsc{OcGCN}.
In the second step, we take a first-order traversal to search the best neighbor nodes of two target objects, respectively, where the best neighbor nodes have the highest connecting scores to their target objects.
Note that we only consider the direct neighbor of target objects (i.e., first-order connection), because including nodes in too distant will rather lead to inaccurate descriptions.
In the third step, we assemble the two perspective structures into one, and then prune the conflicting edge with lower connecting score if a cycle exists (i.e. two target objects connects to a common neighbor), resulting in a successful subgraph.

It is noteworthy that during training we sample only one subgraph with highest-scored edge, where reparameterization trick \cite{BlumHP15} is used for gradient propagation.
During inference, we sample multiple scene subgraphs with top-$k$ highest score edges, i.e., giving diverse descriptions.

With the subgraph at hand, we create its representations via a mean-pooling operation over it:
\setlength\abovedisplayskip{3pt}
\setlength\belowdisplayskip{3pt}
\begin{equation}
\begin{split}
    \lambda_i &= \frac{\delta(v_i\in {G}_{sub})+a_{t_1,i}+a_{t_2,i}}{\sum_{l\in {G}}\left(a_{t_1,l}+a_{t_2,l}\right)}  \,,\\
    \bm{r}^G &= \text{MeanPool}(\{\lambda_i\hat{\bm{s}}^v_i, \lambda_i\hat{\bm{s}}^e_{i,j} | {G}\}) \,,
\end{split}
\label{equation:g}
\end{equation}
where $a_{i,j}$ is the strength score, $t_1$ and $t_2$ are the target nodes, $\delta(exp)$ funtion outputs 1 when $exp$ is true, otherwise 0 , $\hat{\bm{s}}^v_i, \hat{\bm{s}}^e_{i,j}$ are the node and edge representations of last-layer \textsc{OcGCN}.
The local scene graph feature $\bm{r}^G$ will be used during text generation at following stage.

\begin{figure}[t]
    \centering
    \includegraphics[width=0.92\linewidth]{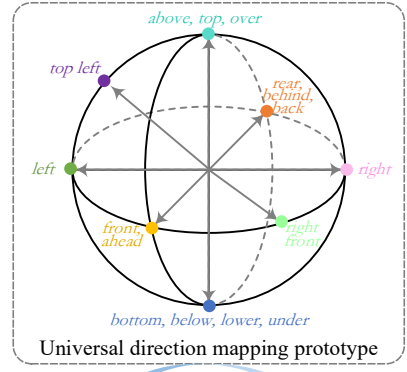}
    \vspace{-1mm}
    \caption{The prototype of direction-term mapping.}
    \label{fig:4}
\end{figure}

\vspace{-1mm}
\paragraph{Building Directional Prompts}

Now we try to guide the VL-PLM to generate contents closely relating to the target objects and the nodes in the sampled subgraph.
We thus build two types of prompt texts, 1) \emph{target object prompt}, e.g.,
\vspace{-1mm}
\begin{center}
    \texttt{<TGT> table <TGT> sofa}
\end{center}
\vspace{-2mm}
and 2) \emph{spatial relation prompt}, e.g.,
\vspace{-1mm}
\begin{center}
    \texttt{<OBJ> table <REL> near <OBJ> sofa} \\
    \texttt{<OBJ> sofa <REL> left <OBJ> bed}
\end{center}
\vspace{-2mm}
Two types of prompts are concatenated as one via a `<SEP>' token.
The former enlightens model what the target objects are, and the latter tells model what possible relational triplets are, i.e., ``<object$_i$, relation, object$_j$>'', where ``relation'' is the predefined relation term.
For each pair of $O_i$, $O_j$ in \textsc{Go3D}-S$^2$G, we map their edge $e_{ij}$ to a specific direction term based on their centroid coordinates.

We maintain a prototype of 3D universal direction-term mapping, as shown in Figure \ref{fig:4}, where we define 26 directions in the whole sphere, with each direction binding certain directional terms.
Even a strong VL-PLM may fail to map a direction to a term accurately.
We thus additionally perform pre-training to strengthen the perception of direction.
\begin{table}[t]
    \fontsize{10}{13}\selectfont
    \setlength{\tabcolsep}{1.mm}
    \begin{center}
    \begin{tabular}{cccccccccc}
    \hline
   \multicolumn{2}{c}{\textbf{Pre-definitions}} \\
    \hline
    \multicolumn{2}{c}{Subject centroid: $x_s$, $y_s$, $z_s$, Object centroid: $x_o$, $y_o$, $z_o$}\\
    \multicolumn{2}{c}{coordinate system: x-toward, y-up, z-right}\\
    \multicolumn{2}{c}{$x,y,z \in \left[0,1\right]$}\\
    \multicolumn{2}{c}{$d_x = |x_s-x_o| \,, d_y = |y_s-y_o| \,, d_z = |z_s-z_o|$}\\
    \hline
    \hline
    \textbf{Rule} & \textbf{Direction Term} \\
    \hline
    \multicolumn{2}{l}{\textbf{Front}: ($d_x > d_y$ and $d_z$, $d_x > 0.2$,$x_s > x_o$)}\\
    $d_y,d_z \leq 0.2$ & ``front'' \\
    $d_y > 0.2, y_s > y_o, d_z \leq 0.2$ & ``front up'' \\
    $d_y > 0.2, y_s < y_o, d_z \leq 0.2$ & ``front down'' \\
    $d_z > 0.2, z_s > z_o, d_y \leq 0.2$ & ``front right'' \\
    $d_z > 0.2, z_s < z_o, d_y \leq 0.2$ & ``front left'' \\
    $d_y,d_z > 0.2, y_s > y_o, z_s > z_o$ & ``front up right'' \\
    $d_y,d_z > 0.2, y_s > y_o, z_s < z_o$ & ``front up left'' \\
    $d_y,d_z > 0.2, y_s < y_o, z_s > z_o$ & ``front down right'' \\
    $d_y,d_z > 0.2, y_s < y_o, z_s < z_o$ & ``front down left'' \\
    \hline
    \multicolumn{2}{l}{\textbf{Back}: ($d_x > d_y$ and $d_z$, $d_x > 0.2$,$x_s < x_o$)}\\
    $d_x > 0.2, d_y,d_z \leq 0.2$ & ``back'' \\
    \multicolumn{2}{c}{\textit{Others are similar to front}}\\
    \hline
    \multicolumn{2}{l}{\textbf{Up}: ($d_y > d_x$ and $d_z$, $d_y > 0.2$,$y_s > y_o$)}\\
    \multicolumn{2}{c}{\textit{Others are similar to front}}\\
    \hline
    \multicolumn{2}{l}{\textbf{Down}: ($d_y > d_x$ and $d_z$, $d_y > 0.2$,$y_s < y_o$)}\\
    \multicolumn{2}{c}{\textit{Others are similar to front}}\\
    \hline
    \multicolumn{2}{l}{\textbf{Right}: ($d_z > d_x$ and $d_y$, $d_z > 0.2$,$z_s > z_o$)}\\
    \multicolumn{2}{c}{\textit{Others are similar to front}}\\
    \hline
    \multicolumn{2}{l}{\textbf{Left}: ($d_z> d_x$ and $d_y$, $d_z > 0.2$,$z_s < z_o$)}\\
    \multicolumn{2}{c}{\textit{Others are similar to front}}\\
    \hline
    ($d_x, d_y, d_z \leq 0.2$) & ``next to''\\
    \hline
    \end{tabular}
    \caption{Direction term mapping rules.}
    \label{tab:direction_mapping}
    \end{center}
    \end{table}
The detailed mapping rules of universal 3D direction-term are shown in Table \ref{tab:direction_mapping}.
With the predefined 26 directions, we compare the centroid coordinates of a pair of objects to decide the direction terms.
Note that according to the rules in Table \ref{tab:direction_mapping}, there may be multiple terms for the same direction, e.g., ``left up front'' and ``up left front'' and we keep these redundant terms in our implementation.
We add an extra term ``next to'' to describe the situation that two objects are close to each other.
For some types of object that not exists during the 3D Scene Extractor pretraining, we just use their 2D locations and treat the depth coordinate to 0.

Moreover, we conduct a pre-training to strengthen the perception
of direction for VL-PTM. Concretely, we utilize the 3D scene extractor and relation triplets ground-truth in VSD dataset to generate a set of pseudo data. 
For example, if we have two target objects $O_1$, $O_2$ with theirs names $Tag_1$, $Tag_2$ ,ground-truth relation term $Rel_g$ (in VSD annotations), and 2D boxes $Box_1$, $Box_2$, we could get their 3D centroid coordinates $loc_1$, $loc_2$ through off-the-shelf 3D scene extractor. 
Then we map the 3D centroid coordinates to 3D direction term $Rel_p$. We use ``\texttt{<OBJ>} $Tag_1$ \texttt{<REL>} $Rel_p$ \texttt{<OBJ>} $Tag_2$'' as inputs and ``$Tag_1$, $Rel_g$, $Tag_2$'' as outputs to train the VL-PTM. Moreover, we randomly replace the $Rel_g$ with some synonyms for data augmentation.

\vspace{-1mm}
\paragraph{Generating Text}

Finally, we feed the prompt text and the raw image as input into our backbone VL-PLM encoder, where the resulting representations $\bm{r}^T$ and $\bm{r}^I$ and the local scene graph feature $\bm{r}^G$ are fused together via cross-attention, i.e., $\bm{r}$=CrossAtt($\bm{r}^G,\bm{r}^T,\bm{r}^I$).
The VL-PLM decoder then performs text generation based on $\bm{r}$.

\begin{table*}[t]
\fontsize{9}{11.5}\selectfont
\setlength{\tabcolsep}{1.3mm}
\begin{center}
\resizebox{0.98\textwidth}{!}{
\begin{tabular}{clcccccccccc}
\hline
\multicolumn{2}{c}{\multirow{2}{*}{}} & \multicolumn{5}{c}{\bf \texttt{VSD}-v1} & \multicolumn{5}{c}{\bf \texttt{VSD}-v2} \\
\cmidrule(r){3-7} \cmidrule(r){8-12}
\multicolumn{2}{c}{} &\bf BLEU-4 & \bf METEOR &\bf ROUGE &\bf CIDEr &\bf SPICE &\bf BLEU-4 &\bf METEOR &\bf ROUGE &\bf CIDEr &\bf SPICE \\
\hline
\multicolumn{12}{l}{$\bullet$ \textbf{VL-PTMs}} \\
& Oscar & 37.17 & 35.06 & 66.47 & 427.21 & 67.41 & 20.90 & 23.83 & 50.96 & 221.61 & 44.12\\
& VL-Bart & 52.71 & 41.96 & 77.57 & 471.21 & 67.83 & 23.78 & 24.83 & 48.49 & 253.26 & 45.04 \\  
& VL-T5 & 52.58 & 41.94 & 77.63 & 472.24 & 67.90 & 23.83 & 24.26 & 53.51 & 255.51 & \underline{46.86} \\  
& OFA & 53.59 & 41.74 & 77.68 & 469.23 & 67.03 & \underline{24.53} & \underline{24.93} & 54.27 & 257.29 & 45.63 \\
\hline
\multicolumn{12}{l}{$\bullet$ \textbf{VL-PTMs + VSRC} \cite{zhaoyu-vsd}} \\
& VLBart-ppl & 53.49 & 42.14 & 77.79 & 474.34 & 67.97 & 24.44 & 24.08 & 53.80 & 256.52 & 45.16 \\
& VLT5-ppl & 53.71 & 42.56 & 78.33 & 480.32 & 68.72 & 23.79 & 24.49 & 54.49 & 256.70 & 46.04 \\
& VLBart-e2e & 53.60 & 42.45 & 78.15 & 476.47 & 68.18 & 24.71 & 24.41 & 54.22 & 258.18 & 45.79 \\
& VLT5-e2e & \underline{54.31} & \underline{42.63} & \underline{78.38} &  \underline{481.13} & \underline{68.74} & 24.47 & 24.50 & \underline{54.52} & \underline{261.70} & 46.07 \\
\hline
\multicolumn{12}{l}{$\bullet$ \textbf{VL-PTMs + 3D scene features}} \\
& \textsc{3Dvsd} (Ours) & \textbf{54.85} & \textbf{43.25} & \textbf{79.38} & \textbf{483.05} & \textbf{68.76} & \textbf{26.40} & \textbf{26.87} & \textbf{55.76} & \textbf{272.93} & \textbf{46.97} \\
\specialrule{0em}{-2pt}{-2pt}& & \scriptsize{(+0.54})  & \scriptsize{(+0.62})  & \scriptsize{(+1.00}) & \scriptsize{(+1.92})  & \scriptsize{(+0.02})  & \scriptsize{(+1.87}) & \scriptsize{(+1.94})  & \scriptsize{(+1.24})  & \scriptsize{(+11.23})  & \scriptsize{(+0.11}) \\
\hline
\end{tabular}
}
\vspace{-1mm}
\caption{
Main results on two datasets.
Bold numbers are the best, and underlined ones are the second best.
}
\label{tab:result}
\end{center}
\end{table*}

\section{Experiments}

\vspace{-1mm}
\subsection{Settings}

\vspace{-1mm}
\paragraph{Dataset}
We evaluate our model on two datasets: VSD-v1 and VSD-v2 \cite{zhaoyu-vsd}.
VSD-v1 is the initial version of VSD datasets, which has a large scale but simple annotations. 
VSD-v2 has the same image source while more complex and human-friendly descriptions, which is more challenging.
We use the original split of train/dev/test set of each dataset.

\vspace{-2mm}
\paragraph{Implementation}
Our model takes the pre-trained 3D extractor from \citet{DBLP:conf/cvpr/NieHGZCZ20}, containing the layout estimation network and the 3D object detection network.
The hidden size of \textsc{OcGCN} is 768 which is the same with text decoder.
The dimension of edge feature $\bm{s}^e_{i,j}$ is also 768.
We adopt the OFA$_{base}$ as our backbone VL-PLM.

\vspace{-2mm}
\paragraph{Evaluation}
We make comparisons with 1) the existing popular image-to-text VL-PLMs, including OSCAR \cite{DBLP:conf/eccv/Li0LZHZWH0WCG20}, VLT5/VLBart \cite{DBLP:conf/icml/ChoLTB21}, OFA \cite{DBLP:conf/icml/WangYMLBLMZZY22};
2) the models introduced in \citet{zhaoyu-vsd}, including the pipeline (ppl) and the end-to-end (e2e) paradigms.
\citet{zhaoyu-vsd} use the visual spatial relations classification (VSRC) results as intermediate features for VSD.
Following \citet{zhaoyu-vsd}, we adopt five automatic metrics to evaluate performances, including BLEU-4, MENTEOR, ROUGE, CIDEr and SPICE.
We measure the diversity with three metrics, i.e., mBLEU-4, BLEU-4@K and SPICE@K.
All the used VL-PLMs are the base version.
Our results are the average scores over five runs.
Appendix $\S$\ref{app:hyperparameter} details all the experimental settings.

\begin{table}[!t]
    \fontsize{9}{11.5}\selectfont
    \setlength{\tabcolsep}{1.4mm}
    \begin{center}
    \begin{tabular}{clcccccccccc}
    \hline
    \multicolumn{2}{c}{} & \bf B4 & \bf M & \bf R & \bf C & \bf S\\
    \hline
    & VL-T5 & 29.41 & 28.67 & 59.23 & 294.81 & 50.04\\
    & OFA & 33.14 & 30.02 & 60.87 & 290.93 & 49.23\\
    & VLT5-e2e & 29.64 & 28.88 & 60.13 & 291.32 & 51.01\\
    \hline
    & \textsc{3Dvsd} & \bf 39.29 & \bf 34.27 & \bf 67.88 & \bf 328.56 & \bf 55.42\\
    \hline
    \end{tabular}
\vspace{-2mm}
    \end{center}
    \caption{Results on hard cases in VSD-v2, where the images come with layout-overlapped and irregularly-posed target objects.
    }
    \label{tab:hard_case}
\end{table}

\subsection{Main Observations}

\begin{table}[t]
\fontsize{9}{11.5}\selectfont
\setlength{\tabcolsep}{1.mm}
\begin{center}
\begin{tabular}{clcc}
\hline
\multicolumn{2}{c}{} & \bf BLEU-4 & \bf SPICE \\
\hline
& \textsc{3Dvsd} (Full) & \bf 26.40 & \bf 46.97 \\
\cdashline{1-4}
& \quad w/o \textsc{Go3D}-S$^2$G & 23.31(-3.09) & 43.89(-3.08) \\
& \quad w/o \textsc{OcGCN} & 24.51(-1.89) & 46.85(-0.12) \\
& \quad w/o S$^3$ mechanism & 25.19(-1.21) & 46.38(-0.59) \\
& \quad w/o Dir-term Prompts & 26.18(-0.22) & 46.17(-0.80) \\
\hline
\end{tabular}
\vspace{-1mm}
\caption{Ablation results (VSD-v2).
`w/o \textsc{Go3D}-S$^2$G' means ablating the graph modeling of 3D scene feature, while instead using the embedded vectors $\bm{s}_i^v$ for text generation.
`w/o \textsc{OcGCN}' means replacing \textsc{OcGCN} encoder with the vanilla GCN.
`\quad w/o S$^3$' means replacing with beam search decoding.
`Dir-term Prompts' represents the directional term prompts.
}
\label{tab:ablation}
\end{center}
\vspace{-3mm}
\end{table}


\paragraph{Main Results}
As shown in Table \ref{tab:result},
overall, the VSD-v2 can be more challenging than VSD-v1, where model scores on all the metrics are lower. 
Also we see that four different VL-PTMs show the similar level of performances, due to their general purpose of pre-training for multimodal learning.
By taking advantages of the VSRC features, \citet{zhaoyu-vsd}'s methods outperform the baseline vanilla VL-PTMs on the task.
However, the improvements from \citet{zhaoyu-vsd}'s models can be incremental, due to the reason that \citet{zhaoyu-vsd} model the input images with only 2D information.
On the contrast, our proposed \textsc{3Dvsd} model achieves significant improvement over the baselines cross two datasets on all the metrics, evidently demonstrating its efficacy.

In addition, our model shows larger improvements on the harder VSD-v2 data than that on VSD-v1.
To directly measure the capability of our method, we further collect a subset from VSD-v2, where the target objects in images are irregularly-posed with complex spatial relation, and also there are overlapped layouts.
We perform experiments on the set, where the results are shown in Table \ref{tab:hard_case}.
We can find that our \textsc{3Dvsd} model improves the best-performing baseline with marked boosts, i.e., 1.87 BLEU-4, 1.94 METROR, 1.24 ROUGE and 5.11 CIDEr.
This significantly indicates that our method is capable of well understanding the visual spatial semantics and thus generating more diverse and flexible VSD sentences.

\paragraph{Model Ablation}

Now we quantify the contribution of each design in our systems via model ablation, as shown in Table \ref{tab:ablation}.
First, we can see that the 3D scene feature from \textsc{Go3D}-S$^2$G graph modeling gives the biggest influences, i.e., contributing 3.09 BLEU-4 and 3.08 SPICE scores.
Besides, the \textsc{OcGCN} encoder, the S$^3$ mechanism as well as the direction-term prompting also plays an essential role to the overall system, respectively.

\paragraph{Evaluation on Spatial-diversity Generation}

As we equip our system with the S$^3$ mechanism, we can generate spatially-diversified texts.
Next, we directly assess the ability on the generation spatial-diversification.
We first make comparisons with the beam search method using automatic metrics \cite{DBLP:conf/cvpr/DeshpandeAWSF19}, including mBLEU-4, BLEU-4@K and SPICE@K.
mBLEU-4 compares the 4-gram matching between one of the generated sentence and the remaining generated sentences for an image, and thus lower mBLEU-4 score means more diversity.
BLEU-4@K and SPICE@K represent the highest BLEU-4 and SPICE score for the top-$k$ generated sentences for an image, where higher BLEU-4@K and SPICE@K prove that the a model can keep better semantics accuracy while generating diverse results.
As shown in Table \ref{tab:diversity}, our S$^3$ mechanism achieves lower mBLEU-4 and higher BLEU-4@K and SPICE@K, demonstrating that our method could generate diversified descriptions with enough semantics accuracy.

\begin{table}[t]
    \fontsize{9}{11.5}\selectfont
    \setlength{\tabcolsep}{1.mm}
    \begin{center}
    \begin{tabular}{clcccccccccc}
    \hline
    \multicolumn{2}{c}{} & \multicolumn{3}{c}{K=5 Samlpling} \\
    \cmidrule(r){3-5}
    \multicolumn{2}{c}{} &\bf  mBLEU-4$\downarrow$ &\bf  BLEU-4@K$\uparrow$ &\bf  SPICE@K$\uparrow$\\
    \hline
    \multicolumn{4}{l}{\textbf{Beam Search}} \\
    & VL-T5 & 8.62 & 33.02 & 60.55 \\
    & OFA & 7.77 & 32.72 & 60.37 \\
    & \textsc{3Dvsd} & 7.6 & 33.12 & 60.67 \\
    \hline
    \multicolumn{4}{l}{\textbf{Scene Subgraph Sampling}} \\ 
    & \textsc{3Dvsd} & 5.01 & 34.44 & 61.99\\
    \hline
    \end{tabular}
\vspace{-1mm}
    \caption{
    Auto-evaluation on spatial diversification.
    }
    \label{tab:diversity}
    \end{center}
\end{table}

\begin{table}[t]
    \fontsize{9}{11.5}\selectfont
    \setlength{\tabcolsep}{1.2mm}
    \begin{center}
    \begin{tabular}{clcccccccccc}
    \hline
    \multicolumn{2}{c}{\multirow{2}{*}{}} & \textbf{Spatial Acc.$\uparrow$} & \textbf{Spatial Div.$\uparrow$} & \textbf{Fluency$\uparrow$}\\
    \hline
    \multicolumn{4}{l}{\textbf{Beam Search}} \\
    & VL-T5 & 3.05 & 2.91 & 4.33 \\
    & OFA & 3.14 & 2.94 & 4.39 \\
    & \textsc{3Dvsd} & 3.13 & 3.10 & 4.49 \\
    \hline
    \multicolumn{4}{l}{\textbf{Scene Subgraph Sampling}} \\ 
    & \textsc{3Dvsd} & 3.19 & 3.98 & 4.53\\
    \hline
    \end{tabular}
\vspace{-1mm}
    \caption{Human evaluation (with Likert 5-scale) on generation diversification.
    We randomly select 100 samples from VSD-v2.
    }
    \label{tab:diversity_human}
    \end{center}
\end{table}

\pgfplotsset{compat=1.7,every axis title/.append style={at={(0.5,-0.45)}, font=\fontsize{14}{1}\selectfont},every axis/.append style={xtick pos=left,ytick pos=left,tickwidth=1.5pt}}
\usetikzlibrary{matrix}
\usepgfplotslibrary{groupplots}
\usetikzlibrary{patterns,backgrounds}

\definecolor{c1}{RGB}{252,232,212}
\definecolor{c2}{RGB}{184,183,163}
\definecolor{c3}{RGB}{107,112,092}
\definecolor{c4}{RGB}{203,153,126}
\definecolor{c5}{RGB}{107,112,092}

\begin{figure}[t]
\centering
\begin{tikzpicture}
\begin{axis}[
	ybar,
	ytick = {22, 24.5, 27, 29.5},
	yticklabels={22, 24.5, 27, 29.5},
	ymax=29.55,ymin=22,
	ylabel = BLEU-4,
	y tick label style = {yshift=-0.5em, text height=0ex,font=\scriptsize},
    x label style = {font=\scriptsize},
    y label style = {yshift=-0.5em, font=\scriptsize},
    axis x line*=bottom,
	axis line style={-},
    axis y line*=left,
	axis line style={-},
	enlargelimits=0.05,
    nodes near coords,
    every node near coord/.append style={black, font=\tiny, opacity=0.7, yshift=-0.0em,xshift=0.0em },
	legend style={at={(0.02,0.99)},anchor=north west, draw=none, legend columns=-1, font=\scriptsize},
	xticklabels={BLEU-4,SPICE},
	xtick={2,6},
	xmax=7.5, xmin=0.5,
	x tick label style = {yshift=0.05em, align=center,font=\small},
	title style={yshift=-0.9em,font=\small},
	width = 7.5cm,
	height = 4cm,
	]
	\addplot[c4, pattern= north east lines, thick, pattern color=c4, bar shift=0pt, bar width = 1.6em] coordinates
	{
		(1.34,  22.91)
	};\addlegendentry{2D Info}
	
	\addplot[c3, pattern=crosshatch, pattern color=c3, thick, bar shift=0pt, bar width = 1.6em] coordinates
	{
		(2.66,  26.40)
	};\addlegendentry{3D Info}
\end{axis}

\begin{axis}[
	ybar,
	ytick = {43, 45, 47},
	yticklabels={43, 45, 47},
	ymax=47.35,ymin=42.9,
	ylabel = SPICE,
	y tick label style = {yshift=-0.5em, text height=0ex,font=\scriptsize},
	x label style = {font=\scriptsize},
    y label style = {yshift=0.5em,font=\scriptsize},    
	axis y line*=right,
	enlargelimits=0.05,
    nodes near coords,
    every node near coord/.append style={black, font=\tiny, opacity=0.7, yshift=-0.em,xshift=0.0em },
	xticklabels={SPICE},
	xmax=7.5, xmin=0.5,
	title style={yshift=-0.9em,font=\small},
	width = 7.5cm,
	height = 4cm,
	axis x line*=top,
	axis line style={-},
	xtick=\empty,
	]
	
	\addplot[c4, pattern= north east lines, thick, pattern color=c4, bar shift=0pt, bar width = 1.6em] coordinates
	{
		(5.34,  43.55)
	};
	\addplot[c3, pattern=crosshatch, pattern color=c3, thick, bar shift=0pt, bar width = 1.6em] coordinates
	{
		(6.66,  46.97)
	};
\end{axis}
\end{tikzpicture}
\vspace{-2mm}
\caption{Comparison of 2D and 3D method on VSDv2.}
\label{fig:3d}
\end{figure}
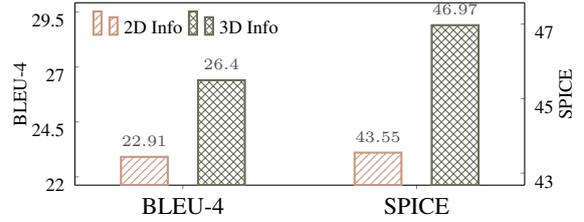

We also provide a human evaluation to describe spatial diversity with respect to the \emph{Spatial Accuracy}, \emph{Spatial Diversity} and \emph{Fluency}.
We ask 10 English speakers to answer the 5-point Likert scale on 100 samples, where the average results are shown in Table \ref{tab:diversity_human}.
Overall, all the models can achieve competitive score on language fluency, thanks to the superiority of VL pretraining.
Also the tendencies of spatial accuracy and diversity is consistent with the automatic evaluation. 
Remarkably, our S$^3$ mechanism shows the best capability on spatial diversity, demonstrating its effectiveness.

\pgfplotsset{compat=1.7,every axis title/.append style={at={(0.5,-0.45)}, font=\fontsize{14}{1}\selectfont},every axis/.append style={xtick pos=left,ytick pos=left,tickwidth=1.5pt}}
\usetikzlibrary{matrix}
\usepgfplotslibrary{groupplots}
\usetikzlibrary{patterns,backgrounds}

\definecolor{c1}{RGB}{252,232,212}
\definecolor{c2}{RGB}{184,183,163}
\definecolor{c3}{RGB}{107,112,092}
\definecolor{c4}{RGB}{203,153,126}
\definecolor{c5}{RGB}{107,112,092}

\begin{figure}[t]
\centering
\begin{tikzpicture}
\begin{axis}[
	ybar,
	ytick = {23.5, 25, 26.6, 28},
	yticklabels={23.5, 25, 26.6, 28},
	ymax=28.05,ymin=23.5,
	ylabel = BLEU-4,
	y tick label style = {yshift=-0.5em, text height=0ex,font=\scriptsize},
    x label style = {font=\scriptsize},
    y label style = {yshift=-0.5em,font=\scriptsize},
	axis line style={-},
    axis x line*=bottom,
    axis y line*=left,
	enlargelimits=0.05,
    nodes near coords,
    every node near coord/.append style={black, font=\tiny, opacity=0.7, yshift=-0.0em,xshift=0.0em },
	legend style={at={(0.02,0.99)},anchor=north west, draw=none, legend columns=-1, font=\scriptsize},
	xticklabels={BLEU-4,SPICE},
	xtick={2,6},
	xmax=7.5, xmin=0.5,
	x tick label style = {yshift=0.05em, align=center,font=\small},
	title style={yshift=-0.9em,font=\small},
	width = 7cm,
	height = 4cm,
	]
	\addplot[c4, pattern= north east lines, thick, pattern color=c4, bar shift=-5pt, bar width = 1em] coordinates
	{
		(1.3,  24.13)
	};\addlegendentry{w/o orientation}

    \addplot[c2, pattern= north east lines, thick, pattern color=c2, bar shift=0pt, bar width = 1em] coordinates
	{
		(2,  24.26)
	};\addlegendentry{w/o size}
	
	\addplot[c3, pattern=crosshatch, pattern color=c3, thick, bar shift=5pt, bar width = 1em] coordinates
	{
		(2.7,  26.40)
	};\addlegendentry{Full}
\end{axis}

\begin{axis}[
		ybar,
	    ytick = {43.5, 45, 46.5, 48},
	    yticklabels={43.5, 45, 46.5, 48},
	    ymax=48.05,ymin=43.5,
        ylabel = SPICE,
	    y tick label style = {yshift=-0.5em, text height=0ex,font=\scriptsize},
        x label style = {yfont=\scriptsize},
        y label style = {yshift=0.5em,font=\scriptsize},
	    axis line style={-},
	    axis y line*=right,
	    enlargelimits=0.05,
    nodes near coords,
    every node near coord/.append style={black, font=\tiny, opacity=0.7, yshift=-0.0em,xshift=0.0em },
	    xmax=7.5, xmin=0.5,
	    x tick label style = {xshift=-5em, yshift=0.05em, align=center,font=\small},
	    title style={yshift=-0.9em,font=\small},
	    width = 7cm,
	    height = 4cm,
        axis x line*=top,
        axis line style={-},
        xtick=\empty,
	]
	
	\addplot[c4, pattern= north east lines, thick, pattern color=c4, bar shift=-5pt, bar width = 1em] coordinates
	{
		(5.3,  44.17)
	};

    \addplot[c2, pattern= north east lines, thick, pattern color=c2, bar shift=0pt, bar width = 1em] coordinates
	{
		(6,  44.25)
	};

	\addplot[c3, pattern=crosshatch, pattern color=c3, thick, bar shift=5pt, bar width = 1em] coordinates
	{
		(6.7,  46.97)
	};
\end{axis}
\end{tikzpicture}
\vspace{-2mm}
\caption{Ablation results of 3D features on VSDv2.}
\label{fig:3d_ablation}
\end{figure}
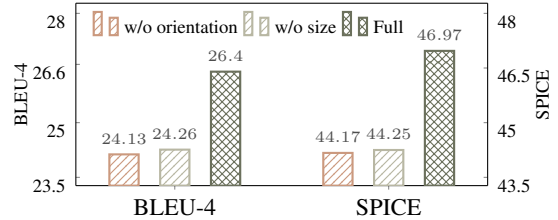

\subsection{Analyses and Discussions}

To gain an in-depth understanding of our method's strengths, we try to answer following research questions via further analyses:

\vspace{5pt}
\noindent\textbf{$\bullet$ RQ1}: \emph{How 3D scene features help understanding the spatial semantics of input images?}

\textbf{A:}
The key of our method is the leverage of 3D scene features.
Now, we first consider downgrading the 3D features into the 2D ones, such that we can gain the perception of its necessity.
To ensure the fair comparison, we remove the 3D scene extractor and replace the 3D pose feature $pose_i$ (Eq. \ref{equation:v}) with the 2D size. 
Also we replace the 3D coordinates by 2D coordinates. 
And the other settings are kept the same.
As shown the results in Figure \ref{fig:3d}, the 2D scene modeling results in the markedly performance decrease.

We can further quantify the contributions of each type of 3D features via feature ablation, including the orientation feature $ori_i$ and the size feature $size_i$.
As seen in Figure \ref{fig:3d_ablation}, both 3D orientation and 3D size features contribute to the overall system.
Also the influence on SPICE is larger, which indicates that the orientation and size of the objects especially help recognize the spatial relation.
Finally, in Figure \ref{fig:case} we empirically show the \textsc{OcGCN}'s kernel weight $\gamma_{i,j}$ (Eq. \ref{OcGCN}) on two pieces of instances, where the attention values reflect the contribution of each object.
It is clear that our model has successfully captured the spatial semantics of the target object pairs.

\vspace{5pt}
\noindent\textbf{$\bullet$ RQ2}: \emph{How does S$^3$ mechanism aid the diversified spatial description generation?}

\input{figures/case_study.tex}

\definecolor{color1}{RGB}{252,232,212}
\definecolor{color2}{RGB}{184,183,163}
\definecolor{color3}{RGB}{107,112,092}
\definecolor{color4}{RGB}{203,153,126}
\definecolor{color5}{RGB}{107,112,092}

\makeatletter

\tikzstyle{chart}=[
    legend label/.style={font={\scriptsize},anchor=west,align=left},
    legend box/.style={rectangle, draw, minimum size=5pt},
    axis/.style={black,semithick,->},
    axis label/.style={anchor=east,font={\tiny}},
]

\tikzstyle{bar chart}=[
    chart,
    bar width/.code={
        \pgfmathparse{##1/2}
        \global\let\bar@w\pgfmathresult
    },
    bar/.style={very thick, draw=white},
    bar label/.style={font={\bf\small},anchor=north},
    bar value/.style={font={\footnotesize}},
    bar width=.75,
]

\tikzstyle{pie chart}=[
    chart,
    slice/.style={line cap=round, line join=round, very thick,draw=white},
    pie title/.style={font={\bf}},
    slice type/.style 2 args={
        ##1/.style={fill=##2},
        values of ##1/.style={}
    }
]

\pgfdeclarelayer{background}
\pgfdeclarelayer{foreground}
\pgfsetlayers{background,main,foreground}

\newcommand{\pie}[3][]{
    \begin{scope}[#1]
    \pgfmathsetmacro{\curA}{90}
    \pgfmathsetmacro{\r}{1}
    \def\c{(0,0)}
    \node[pie title] at (90:1.3) {#2};
    \foreach \v/\s in{#3}{
        \pgfmathsetmacro{\deltaA}{\v/100*360}
        \pgfmathsetmacro{\nextA}{\curA + \deltaA}
        \pgfmathsetmacro{\midA}{(\curA+\nextA)/2}

        \path[slice,\s] \c
            -- +(\curA:\r)
            arc (\curA:\nextA:\r)
            -- cycle;
        \pgfmathsetmacro{\d}{max((\deltaA * -(.5/50) + 1) , .5)}

        \begin{pgfonlayer}{foreground}
        \path \c -- node[pos=\d,pie values,values of \s]{$\v\%$} +(\midA:\r);
        \end{pgfonlayer}

        \global\let\curA\nextA
    }

    \draw (0,-1) -- (-0.05,-1.2) -- (-0.2,-1.2);
    \node () at (-0.4,-1.2) {\small{VG}};

    \draw (60:1) -- (0.6,1) -- (0.7,1);
    \node () at (1,1) {\small{Flickr}};

    \draw (15:1) -- (1,0.4);
    \node () at (1.2,0.5) {\small{NYU}};

    \end{scope}
}

\newcommand{\legend}[2][]{
    \begin{scope}[#1]
    \path
        \foreach \n/\s in {#2}
            {
                  ++(0,-10pt) node[\s,legend box] {} +(5pt,0) node[legend label] {\n}
            }
    ;
    \end{scope}
}

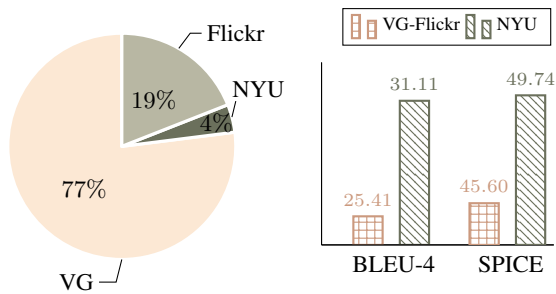
\begin{figure}
    \begin{minipage}{.5\linewidth}
        \centering
                \begin{tikzpicture}[
                    pie chart,
                    slice type={vg}{color1},
                    slice type={flickr}{color2},
                    slice type={nyu}{color3},
                    pie values/.style={font={\small}},
                    scale=1.5
                ]
                    
                \pie{}{77/vg,4/nyu,19/flickr}

                \end{tikzpicture}
    \end{minipage}%
    \begin{minipage}{.5\linewidth}
        \centering
            \pgfkeys{/pgf/number format/.cd,fixed, fixed zerofill,precision=2}
            \begin{tikzpicture}
            \begin{axis}[
                ybar,
                bar width=10pt,
                xtick = \empty,
                xmax=5,xmin=0,
                x tick label style = {yshift=-0.5em, text height=0ex,font=\scriptsize},
                ytick = \empty,
                ymax=33,ymin=24,
                y tick label style = {yshift=-0.3em, text height=0ex,font=\scriptsize},
                x label style = {font=\scriptsize},
                y label style = {font=\scriptsize},
                axis x line = bottom,
                axis y line=left,
                height = 4cm,
                axis line style={-},
                title style={xshift=-1.5em,yshift=1em,font=\small},
                title=BLEU-4,
                nodes near coords style={font=\scriptsize},
                nodes near coords,
                legend style={xshift=-3.5em, yshift=2em, text height=0ex, anchor=north, legend columns=2, font=\scriptsize},
            ]
            \addplot[color4, pattern=grid, thick, pattern color=color4, bar shift=0pt, bar width = 1em] coordinates
            {
                (1,  25.41)
            };\addlegendentry{VG-Flickr}
            \addplot[color3, pattern=north west lines, pattern color=color3, thick, bar shift=0pt, bar width = 1em] coordinates
            {
                (2,  31.11)
            };\addlegendentry{NYU}
            \end{axis}
            \begin{axis}[
                ybar,
                bar width=10pt,
                xtick = \empty,
                xmax=5,xmin=0,
                x tick label style = {yshift=-0.5em, text height=0ex,font=\scriptsize},
                ytick = \empty,
                ymax=51,ymin=44,
                y tick label style = {yshift=-0.3em, text height=0ex,font=\scriptsize},
                x label style = {font=\scriptsize},
                y label style = {font=\scriptsize},
                axis x line = bottom,
                axis y line=left,
                height = 4cm,
                axis line style={-},
                title style={xshift=2.5em,yshift=1em,font=\small},
                title=SPICE,
                nodes near coords style={font=\scriptsize},
                nodes near coords,
                legend style={xshift=-3em, yshift=2.5em, text height=0ex, anchor=north, legend columns=2, font=\scriptsize},
            ]
            \addplot[color4, pattern=grid, thick, pattern color=color4, bar shift=0pt, bar width = 1em] coordinates
            {
                (3.5,  45.60)
            };
            \addplot[color3, pattern=north west lines, thick, pattern color=color3, bar shift=0pt, bar width = 1em] coordinates
            {
                (4.5,  49.74)
            };
            \end{axis}
            \end{tikzpicture}
    \end{minipage}%
\vspace{-1mm}
\caption{
Performances on the indoor (NYU) and outdoor (Visual Genome and Flickr) input images.
The pie chart shows the data proportion.
}
\vspace{-2mm}
\label{fig:indoor}
\end{figure}

\textbf{A:}
Next, we consider investigating how exactly the S$^3$ mechanism contribute to the spatial description generation.
Through our S$^3$ mechanism, we could generate 4 types of subgraphs: 
1) with only target nodes (2-hop); 
2) with one non-target neighbor node linked with subject node (3-hop-s); 
3) with one non-target neighbor node linked with object node (3-hop-o); 
4) with two non-target neighbor nodes. 
In our implementation, we use a threshold $p_{cut}$ to filter out edges with very low scores. 
Figure \ref{fig:hops} show the distribution of four subgraphs with $p_{cut}$=0.1 or 0.2. 
We see that different $p_{cut}$ values help generates spatial descriptions with varying numbers of objects attended into the subgraphs.
That is, S$^3$ mechanism aids the diversified spatial description generation by producing multiple heterogeneous subgraphs structures.


\definecolor{color1}{RGB}{252,232,212}
\definecolor{color2}{RGB}{184,183,163}
\definecolor{color3}{RGB}{107,112,092}
\definecolor{color4}{RGB}{203,153,126}
\definecolor{color5}{RGB}{107,112,092}

\makeatletter

\tikzstyle{chart}=[
    legend label/.style={font={\scriptsize},anchor=west,align=left},
    legend box/.style={rectangle, draw, minimum size=5pt},
    axis/.style={black,semithick,->},
    axis label/.style={anchor=east,font={\tiny}},
]

\tikzstyle{bar chart}=[
    chart,
    bar width/.code={
        \pgfmathparse{##1/2}
        \global\let\bar@w\pgfmathresult
    },
    bar/.style={very thick, draw=white},
    bar label/.style={font={\bf\small},anchor=north},
    bar value/.style={font={\footnotesize}},
    bar width=.75,
]

\tikzstyle{pie chart}=[
    chart,
    slice/.style={line cap=round, line join=round, very thick,draw=white},
    pie title/.style={font={\bf}},
    slice type/.style 2 args={
        ##1/.style={fill=##2},
        values of ##1/.style={}
    }
]

\pgfdeclarelayer{background}
\pgfdeclarelayer{foreground}
\pgfsetlayers{background,main,foreground}

\newcommand{\piehop}[3][]{
    \begin{scope}[#1]
    \pgfmathsetmacro{\curA}{90}
    \pgfmathsetmacro{\r}{1}
    \def\c{(0,0)}
    \node[pie title] at (90:1.3) {#2};
    \foreach \v/\s in{#3}{
        \pgfmathsetmacro{\deltaA}{\v/100*360}
        \pgfmathsetmacro{\nextA}{\curA + \deltaA}
        \pgfmathsetmacro{\midA}{(\curA+\nextA)/2}

        \path[slice,\s] \c
            -- +(\curA:\r)
            arc (\curA:\nextA:\r)
            -- cycle;
        \pgfmathsetmacro{\d}{max((\deltaA * -(.5/50) + 1) , .5)}

        \begin{pgfonlayer}{foreground}
        \path \c -- node[pos=\d,pie values,values of \s]{$\v\%$} +(\midA:\r);
        \end{pgfonlayer}

        \global\let\curA\nextA
    }

    \draw (0,-1) -- (-0.05,-1.2) -- (-0.2,-1.2);
    \node () at (-0.5,-1.2) {\scriptsize{2-hop}};

    \draw (60:1) -- (0.6,1) -- (0.7,1);
    \node () at (1,1) {\scriptsize{3-hop-s}};

    \draw (16:1) -- (1,0.4);
    \node () at (1.2,0.5) {\scriptsize{3-hop-o}};

    \draw (5:1.1) -- (1.2,0);
    \node () at (1.2,-0.1) {\scriptsize{4-hop}};

    \end{scope}
}

\newcommand{\piehoptwo}[3][]{
    \begin{scope}[#1]
    \pgfmathsetmacro{\curA}{90}
    \pgfmathsetmacro{\r}{1}
    \def\c{(0,0)}
    \node[pie title] at (90:1.3) {#2};
    \foreach \v/\s in{#3}{
        \pgfmathsetmacro{\deltaA}{\v/100*360}
        \pgfmathsetmacro{\nextA}{\curA + \deltaA}
        \pgfmathsetmacro{\midA}{(\curA+\nextA)/2}

        \path[slice,\s] \c
            -- +(\curA:\r)
            arc (\curA:\nextA:\r)
            -- cycle;
        \pgfmathsetmacro{\d}{max((\deltaA * -(.5/50) + 1) , .5)}

        \begin{pgfonlayer}{foreground}
        \path \c -- node[pos=\d,pie values,values of \s]{$\v\%$} +(\midA:\r);
        \end{pgfonlayer}

        \global\let\curA\nextA
    }

    \draw (100:1) -- (-0.2,1.1) -- (-0.35,1.1);
    \node () at (-0.6,1.1) {\scriptsize{2-hop}};

    \draw (60:1) -- (0.6,1) -- (0.7,1);
    \node () at (1,1) {\scriptsize{3-hop-s}};

    \draw (-16:1) -- (1,-0.4);
    \node () at (1.15,-0.5) {\scriptsize{3-hop-o}};

    \draw (0,-1) -- (-0.05,-1.2) -- (-0.2,-1.2);
    \node () at (-0.5,-1.2) {\scriptsize{4-hop}};

    \end{scope}
}

\begin{figure}[!t]
    \begin{minipage}{.5\linewidth}
        \centering
                \begin{tikzpicture}[
                    pie chart,
                    slice type={so}{color1},
                    slice type={soaj}{color2},
                    slice type={oaj}{color3},
                    slice type={saj}{color4},
                    pie values/.style={font={\scriptsize}},
                    scale=1.5
                ]
                \piehop{}{76/so,1/soaj,10/oaj,13/saj}
                \end{tikzpicture}
                \caption*{(a) $p_{cut}=0.2$}
    \end{minipage}%
    \begin{minipage}{.5\linewidth}
        \centering
            \begin{tikzpicture}[
                pie chart,
                slice type={so}{color1},
                slice type={soaj}{color2},
                slice type={oaj}{color3},
                slice type={saj}{color4},
                pie values/.style={font={\scriptsize}},
                scale=1.5
            ]
            \piehoptwo{}{45/so,9/soaj,31/oaj,15/saj}
            \end{tikzpicture}
            \caption*{(b) $p_{cut}=0.1$}
    \end{minipage}%
\vspace{-1mm}
\caption{Distribution of subgraph types.}
\label{fig:hops}
\vspace{-3mm}
\end{figure}

We also empirically show the qualitative results in Figure \ref{fig:case}.
We notice that the beam search method can generate multiple alternative texts in both cases, where unfortunately the diversification on describing the spatial relation is much inferior and limited.
In contrast, with our S$^3$ method, the system generates spatial-diversified descriptions for both two images.
Some surrounding objects, e.g., `shelf', `table', `door' in the first case and `chair', `bed', `shelf' and `desk' in the second case, are leveraged to aid describe the target objects.

\vspace{5pt}
\noindent\textbf{$\bullet$ RQ3}: \emph{To what extent the external 3D scene extractor quality influence the VSD performances?}

\textbf{A:}
As we obtain the initial 3D scene features from the external extractor, the quality of the extractor is key to our final VSD performance.
Note that the off-the-shelf 3D extractor is trained on a datasets of indoor scenes, while in VSD dataset the images contain both types of indoor (NYU) and outdoor (Visual Genome (VG) and Flickr).
As shown in Figure \ref{fig:indoor}, the indoor images actually are the minority in our VSD data.
Here we perform analysis to valid the influence on 3D scene extractor.
We split the VSD-v2 test set into indoor and outdoor subsets according to the domain types.
Then we run our model on the two sets separately.
As shown in the figure, the results on the indoor NYU set exceed those on the outdoor VG\&Flickr set clearly, demonstrating the domain shift issue in our system.
Given that our system has already secured considerable performance increase than existing models, we presume that when obtaining a 3D extractor capable of detecting higher-quality 3D scene features for any domain and scenario, our method has the greater potential to gain more task improvements.

\section{Related Work}

\vspace{-1mm}
Image-to-text (I2T) is a fundamental task category of the vision-language multimodal topic.
Existing I2T tasks, e.g., image captioning \cite{DBLP:conf/cvpr/VinyalsTBE15, DBLP:conf/cvpr/CorniaBC19, DBLP:conf/cvpr/MathewsXH18} and VQA \cite{DBLP:conf/iccv/AntolALMBZP15, DBLP:conf/tencon/LubnaKL19, DBLP:journals/air/ManmadhanK20}, attempt to generate textual pieces to understand the image content semantics through different perspectives.
VSD is also a subtask of I2T, which however places the focus on the spatial semantics understanding.
Within recent years, VL-PTMs are extensively employed for the I2T tasks, which have helped achieve state-of-the-art performances on many benchmarks \cite{DBLP:conf/nips/LuBPL19, DBLP:journals/corr/abs-1909-11740,DBLP:conf/aaai/ZhouPZHCG20, DBLP:conf/eccv/Li0LZHZWH0WCG20,DBLP:conf/icml/WangYMLBLMZZY22}.
Prior VSD study  \cite{zhaoyu-vsd} has benchmarked the VSD task with these general-purpose VL-PTMs.
Unfortunately, different from the content-semantic I2T tasks, the goal of VSD is to grasp the spatial semantics.

This work also relates to the theme of visual spatial understanding, which has been intensively investigated in the past years.
\citet{DBLP:conf/iccv/YangRD19, DBLP:journals/tacl/JannerNB18} propose the visual spatial relation classification (VSRC) task,
aiming to predict whether a spatial description is reasonable for the input image.
Differently, VSD aims to directly generate the descriptions of the spatial relations.
The bottleneck of visual spatial understanding tasks is the capturing of spatial semantics. 
Thus, in this paper, we propose improving the visual spatial understanding for VSD via modeling the holistic 3D scene features.
Our work takes the advancements from the topic of 3D scene parsing \cite{HuangQXZWZ18,HuangQZXXZ18,DBLP:conf/cvpr/NieHGZCZ20,ZhangC0ZPL21}, which reconstructs the 3D scene from a single RGB image.
By using the 3D scene features, e.g., 3D layout and 3D object, we achieve the goal of more in-depth understanding of the spatial semantics.

\section{Conclusion}

\vspace{-1mm}
In this work we incorporate the 3D scene features for improving the visual spatial description (VSD) task.
We first employ an off-the-shelf 3D scene extractor to produce 3D objects and scene features for the input images.
We then build a target object-centered 3D spatial scene graph (\textsc{Go3D}-S$^2$G) structure, which is encoded with an object-centered graph convolutional network.
Next, we devise a scene subgraph selecting mechanism to sample topologically-diverse subgraphs from \textsc{Go3D}-S$^2$G, where the local features are used to guide to generate spatially-diversified descriptions.
On the VSD datasets our framework shows superiority on the task, especially for solving the complex cases, such as layout-overlapped and irregularly-posed object.
Meanwhile our method can produce more spatially-diversified generation.
Finally, we demonstrate the influence of quality of the external 3D scene extractor.

\section*{Acknowledgments}

This research is supported by the National Natural Science Foundation of China under Grant No. 62176180.
The work is also supported by the Sea-NExT Joint Lab.

\section*{Limitations}

This work has one major risk.
As the main idea proposed in this work heavily relies on the external 3D scene extractor, the quality of extractor on our used VSD images largely influences the task performance.
However, we reveal in analysis that although suffering from the domain shift issue by the out-of-domain 3D scene extractor, our method still improves the VSD task.
We show that when handling the in-domain VSD images as used for training the 3D scene extractor, the VSD performance has been boosted remarkedly.
Thus, with better a 3D scene extractor, it can be expected that our system will exhibit much stronger capability and advance the VSD task more significantly.

\bibliography{custom}

\begin{thebibliography}{36}
\expandafter\ifx\csname natexlab\endcsname\relax\def\natexlab#1{#1}\fi

\bibitem[{Antol et~al.(2015)Antol, Agrawal, Lu, Mitchell, Batra, Zitnick, and
  Parikh}]{DBLP:conf/iccv/AntolALMBZP15}
Stanislaw Antol, Aishwarya Agrawal, Jiasen Lu, Margaret Mitchell, Dhruv Batra,
  C.~Lawrence Zitnick, and Devi Parikh. 2015.
\newblock {VQA:} visual question answering.
\newblock In \emph{Proceedings of the {ICCV}}, pages 2425--2433.

\bibitem[{Blum et~al.(2015)Blum, Haghtalab, and Procaccia}]{BlumHP15}
Avrim Blum, Nika Haghtalab, and Ariel~D. Procaccia. 2015.
\newblock Variational dropout and the local reparameterization trick.
\newblock In \emph{Proceedings of the {NIPS}}, pages 2575--2583.

\bibitem[{Castaman et~al.(2021)Castaman, Tosello, Antonello, Bagarello, Gandin,
  Carraro, Munaro, Bortoletto, Ghidoni, Menegatti, and
  Pagello}]{CastamanTABGCMB21}
Nicola Castaman, Elisa Tosello, Morris Antonello, Nicola Bagarello, Silvia
  Gandin, Marco Carraro, Matteo Munaro, Roberto Bortoletto, Stefano Ghidoni,
  Emanuele Menegatti, and Enrico Pagello. 2021.
\newblock {RUR53:} an unmanned ground vehicle for navigation, recognition, and
  manipulation.
\newblock \emph{Advanced Robotics}, 35(1):1--18.

\bibitem[{Chen et~al.(2021)Chen, Jiang, Xiao, and Liu}]{0016J0021}
Long Chen, Zhihong Jiang, Jun Xiao, and Wei Liu. 2021.
\newblock Human-like controllable image captioning with verb-specific semantic
  roles.
\newblock In \emph{Proceedings of the {CVPR}}, pages 16846--16856.

\bibitem[{Chen et~al.(2019)Chen, Li, Yu, Kholy, Ahmed, Gan, Cheng, and
  Liu}]{DBLP:journals/corr/abs-1909-11740}
Yen{-}Chun Chen, Linjie Li, Licheng Yu, Ahmed~El Kholy, Faisal Ahmed, Zhe Gan,
  Yu~Cheng, and Jingjing Liu. 2019.
\newblock \href {http://arxiv.org/abs/1909.11740} {{UNITER:} learning universal
  image-text representations}.
\newblock \emph{CoRR}, abs/1909.11740.

\bibitem[{Cho et~al.(2021)Cho, Lei, Tan, and Bansal}]{DBLP:conf/icml/ChoLTB21}
Jaemin Cho, Jie Lei, Hao Tan, and Mohit Bansal. 2021.
\newblock Unifying vision-and-language tasks via text generation.
\newblock In \emph{Proceedings of the {ICML}}, pages 1931--1942.

\bibitem[{Cornia et~al.(2019)Cornia, Baraldi, and
  Cucchiara}]{DBLP:conf/cvpr/CorniaBC19}
Marcella Cornia, Lorenzo Baraldi, and Rita Cucchiara. 2019.
\newblock Show, control and tell: {A} framework for generating controllable and
  grounded captions.
\newblock In \emph{Proceedings of the {CVPR}}, pages 8307--8316.

\bibitem[{Deshpande et~al.(2019)Deshpande, Aneja, Wang, Schwing, and
  Forsyth}]{DBLP:conf/cvpr/DeshpandeAWSF19}
Aditya Deshpande, Jyoti Aneja, Liwei Wang, Alexander~G. Schwing, and David~A.
  Forsyth. 2019.
\newblock Fast, diverse and accurate image captioning guided by part-of-speech.
\newblock In \emph{{IEEE} Conference on Computer Vision and Pattern
  Recognition, {CVPR} 2019, Long Beach, CA, USA, June 16-20, 2019}, pages
  10695--10704. Computer Vision Foundation / {IEEE}.

\bibitem[{Heuser et~al.(2020)Heuser, Arend, and Sunnen}]{HeuserAS20}
Svenja Heuser, B{\'{e}}atrice Arend, and Patrick Sunnen. 2020.
\newblock Reading aloud in human-computer interaction: How spatial distribution
  of digital text units at an interactive tabletop contributes to the
  participants' shared understanding.
\newblock In \emph{Proceedings of the {HCI}}, pages 117--134.

\bibitem[{Huang et~al.(2018{\natexlab{a}})Huang, Qi, Xiao, Zhu, Wu, and
  Zhu}]{HuangQXZWZ18}
Siyuan Huang, Siyuan Qi, Yinxue Xiao, Yixin Zhu, Ying~Nian Wu, and Song{-}Chun
  Zhu. 2018{\natexlab{a}}.
\newblock Cooperative holistic scene understanding: Unifying 3d object, layout,
  and camera pose estimation.
\newblock In \emph{Proceedings of the NeruIPS}, pages 206--217.

\bibitem[{Huang et~al.(2018{\natexlab{b}})Huang, Qi, Zhu, Xiao, Xu, and
  Zhu}]{HuangQZXXZ18}
Siyuan Huang, Siyuan Qi, Yixin Zhu, Yinxue Xiao, Yuanlu Xu, and Song{-}Chun
  Zhu. 2018{\natexlab{b}}.
\newblock Holistic 3d scene parsing and reconstruction from a single {RGB}
  image.
\newblock In \emph{Proceedings of the {ECCV}}, pages 194--211.

\bibitem[{J{\"{a}}nner et~al.(2018)J{\"{a}}nner, Narasimhan, and
  Barzilay}]{DBLP:journals/tacl/JannerNB18}
Michaela J{\"{a}}nner, Karthik Narasimhan, and Regina Barzilay. 2018.
\newblock Representation learning for grounded spatial reasoning.
\newblock \emph{Trans. Assoc. Comput. Linguistics}, 6:49--61.

\bibitem[{Krishna et~al.(2017)Krishna, Zhu, Groth, Johnson, Hata, Kravitz,
  Chen, Kalantidis, Li, Shamma, Bernstein, and Fei{-}Fei}]{KrishnaZGJHKCKL17}
Ranjay Krishna, Yuke Zhu, Oliver Groth, Justin Johnson, Kenji Hata, Joshua
  Kravitz, Stephanie Chen, Yannis Kalantidis, Li{-}Jia Li, David~A. Shamma,
  Michael~S. Bernstein, and Li~Fei{-}Fei. 2017.
\newblock Visual genome: Connecting language and vision using crowdsourced
  dense image annotations.
\newblock \emph{International Journal of Computer Vision}, 123(1):32--73.

\bibitem[{Li et~al.(2020)Li, Yin, Li, Zhang, Hu, Zhang, Wang, Hu, Dong, Wei,
  Choi, and Gao}]{DBLP:conf/eccv/Li0LZHZWH0WCG20}
Xiujun Li, Xi~Yin, Chunyuan Li, Pengchuan Zhang, Xiaowei Hu, Lei Zhang, Lijuan
  Wang, Houdong Hu, Li~Dong, Furu Wei, Yejin Choi, and Jianfeng Gao. 2020.
\newblock Oscar: Object-semantics aligned pre-training for vision-language
  tasks.
\newblock In \emph{Proceedings of the {ECCV}}, pages 121--137.

\bibitem[{Lu et~al.(2019)Lu, Batra, Parikh, and Lee}]{DBLP:conf/nips/LuBPL19}
Jiasen Lu, Dhruv Batra, Devi Parikh, and Stefan Lee. 2019.
\newblock Vilbert: Pretraining task-agnostic visiolinguistic representations
  for vision-and-language tasks.
\newblock In \emph{Proceedings of the {NIPS}}, pages 13--23.

\bibitem[{Lubna et~al.(2019)Lubna, Kalady, and
  Lijiya}]{DBLP:conf/tencon/LubnaKL19}
A.~Lubna, Saidalavi Kalady, and A.~Lijiya. 2019.
\newblock Mobvqa: {A} modality based medical image visual question answering
  system.
\newblock In \emph{Proceedings of the {TENCON}}, pages 727--732.

\bibitem[{Manmadhan and Kovoor(2020)}]{DBLP:journals/air/ManmadhanK20}
Sruthy Manmadhan and Binsu~C. Kovoor. 2020.
\newblock Visual question answering: a state-of-the-art review.
\newblock \emph{Artif. Intell. Rev.}, 53(8):5705--5745.

\bibitem[{Marcheggiani and Titov(2017)}]{marcheggiani-titov-2017-encoding}
Diego Marcheggiani and Ivan Titov. 2017.
\newblock Encoding sentences with graph convolutional networks for semantic
  role labeling.
\newblock In \emph{Proceedings of the EMNLP}, pages 1506--1515.

\bibitem[{Mathews et~al.(2018)Mathews, Xie, and
  He}]{DBLP:conf/cvpr/MathewsXH18}
Alexander~Patrick Mathews, Lexing Xie, and Xuming He. 2018.
\newblock Semstyle: Learning to generate stylised image captions using
  unaligned text.
\newblock In \emph{Proceedings of the {CVPR}}, pages 8591--8600.

\bibitem[{Nie et~al.(2020)Nie, Han, Guo, Zheng, Chang, and
  Zhang}]{DBLP:conf/cvpr/NieHGZCZ20}
Yinyu Nie, Xiaoguang Han, Shihui Guo, Yujian Zheng, Jian Chang, and Jian{-}Jun
  Zhang. 2020.
\newblock Total3dunderstanding: Joint layout, object pose and mesh
  reconstruction for indoor scenes from a single image.
\newblock In \emph{Proceedings of the {CVPR}}, pages 52--61.

\bibitem[{Pend{\~{a}}o and Moreira(2021)}]{PendaoM21}
Cristiano~G. Pend{\~{a}}o and Adriano J.~C. Moreira. 2021.
\newblock Fastgraph enhanced: High accuracy automatic indoor navigation and
  mapping.
\newblock \emph{IEEE Transactions on Mobile Computing}, 20(3):1027--1045.

\bibitem[{Ren et~al.(2015)Ren, He, Girshick, and Sun}]{DBLP:conf/nips/RenHGS15}
Shaoqing Ren, Kaiming He, Ross~B. Girshick, and Jian Sun. 2015.
\newblock Faster {R-CNN:} towards real-time object detection with region
  proposal networks.
\newblock In \emph{Proceedings of the {NIPS}}, pages 91--99.

\bibitem[{Silberman et~al.(2012)Silberman, Hoiem, Kohli, and
  Fergus}]{DBLP:conf/eccv/SilbermanHKF12}
Nathan Silberman, Derek Hoiem, Pushmeet Kohli, and Rob Fergus. 2012.
\newblock Indoor segmentation and support inference from {RGBD} images.
\newblock In \emph{Proceedings of the {ECCV}}, pages 746--760. Springer.

\bibitem[{Song et~al.(2015)Song, Lichtenberg, and
  Xiao}]{DBLP:conf/cvpr/SongLX15}
Shuran Song, Samuel~P. Lichtenberg, and Jianxiong Xiao. 2015.
\newblock {SUN} {RGB-D:} {A} {RGB-D} scene understanding benchmark suite.
\newblock In \emph{Proceedings of the {CVPR}}, pages 567--576.

\bibitem[{Vanhooydonck et~al.(2010)Vanhooydonck, Demeester, H{\"{u}}ntemann,
  Philips, Vanacker, Brussel, and Nuttin}]{VanhooydonckDHPVBN10}
Dirk Vanhooydonck, Eric Demeester, Alexander H{\"{u}}ntemann, Johan Philips,
  Gerolf Vanacker, Hendrik~Van Brussel, and Marnix Nuttin. 2010.
\newblock Adaptable navigational assistance for intelligent wheelchairs by
  means of an implicit personalized user model.
\newblock \emph{Robotics and Autonomous Systems}, 58(8):963--977.

\bibitem[{Vijayakumar et~al.(2018)Vijayakumar, Cogswell, Selvaraju, Sun, Lee,
  Crandall, and Batra}]{VijayakumarCSSL18}
Ashwin~K. Vijayakumar, Michael Cogswell, Ramprasaath~R. Selvaraju, Qing Sun,
  Stefan Lee, David~J. Crandall, and Dhruv Batra. 2018.
\newblock Diverse beam search for improved description of complex scenes.
\newblock In \emph{Proceedings of the {AAAI}}, pages 7371--7379.

\bibitem[{Vinyals et~al.(2015)Vinyals, Toshev, Bengio, and
  Erhan}]{DBLP:conf/cvpr/VinyalsTBE15}
Oriol Vinyals, Alexander Toshev, Samy Bengio, and Dumitru Erhan. 2015.
\newblock Show and tell: {A} neural image caption generator.
\newblock In \emph{Proceedings of the {CVPR}}, pages 3156--3164.

\bibitem[{Wang et~al.(2023{\natexlab{a}})Wang, Luo, Pei, Liu, Huang, Zhang, and
  Yang}]{wang2023entropy}
Chenwei Wang, Siyi Luo, Jifang Pei, Xiaoyu Liu, Yulin Huang, Yin Zhang, and
  Jianyu Yang. 2023{\natexlab{a}}.
\newblock An entropy-awareness meta-learning method for sar open-set atr.
\newblock \emph{IEEE Geoscience and Remote Sensing Letters}.

\bibitem[{Wang et~al.(2023{\natexlab{b}})Wang, Pei, Luo, Huo, Huang, Zhang, and
  Yang}]{wang2023sar}
Chenwei Wang, Jifang Pei, Siyi Luo, Weibo Huo, Yulin Huang, Yin Zhang, and
  Jianyu Yang. 2023{\natexlab{b}}.
\newblock Sar ship target recognition via multiscale feature attention and
  adaptive-weighed classifier.
\newblock \emph{IEEE Geoscience and Remote Sensing Letters}, 20:1--5.

\bibitem[{Wang et~al.(2022{\natexlab{a}})Wang, Pei, Yang, Liu, Huang, and
  Mao}]{wang2022recognition}
Chenwei Wang, Jifang Pei, Jianyu Yang, Xiaoyu Liu, Yulin Huang, and Deqing Mao.
  2022{\natexlab{a}}.
\newblock Recognition in label and discrimination in feature: A hierarchically
  designed lightweight method for limited data in sar atr.
\newblock \emph{IEEE Transactions on Geoscience and Remote Sensing}, 60:1--13.

\bibitem[{Wang et~al.(2022{\natexlab{b}})Wang, Yang, Men, Lin, Bai, Li, Ma,
  Zhou, Zhou, and Yang}]{DBLP:conf/icml/WangYMLBLMZZY22}
Peng Wang, An~Yang, Rui Men, Junyang Lin, Shuai Bai, Zhikang Li, Jianxin Ma,
  Chang Zhou, Jingren Zhou, and Hongxia Yang. 2022{\natexlab{b}}.
\newblock {OFA:} unifying architectures, tasks, and modalities through a simple
  sequence-to-sequence learning framework.
\newblock In \emph{Proceedings of the {ICML}}, volume 162, pages 23318--23340.

\bibitem[{Yang et~al.(2019)Yang, Russakovsky, and
  Deng}]{DBLP:conf/iccv/YangRD19}
Kaiyu Yang, Olga Russakovsky, and Jia Deng. 2019.
\newblock Spatialsense: An adversarially crowdsourced benchmark for spatial
  relation recognition.
\newblock In \emph{Proceedings of the {ICCV}}, pages 2051--2060.

\bibitem[{Yu et~al.(2022)Yu, Zhu, Qin, Zhang, Zhao, and
  Jiang}]{yu-etal-2022-diversifying}
Wenhao Yu, Chenguang Zhu, Lianhui Qin, Zhihan Zhang, Tong Zhao, and Meng Jiang.
  2022.
\newblock Diversifying content generation for commonsense reasoning with
  mixture of knowledge graph experts.
\newblock In \emph{Proceedings of the ACL}, pages 1896--1906.

\bibitem[{Zhang et~al.(2021)Zhang, Cui, Zhang, Zeng, Pollefeys, and
  Liu}]{ZhangC0ZPL21}
Cheng Zhang, Zhaopeng Cui, Yinda Zhang, Bing Zeng, Marc Pollefeys, and
  Shuaicheng Liu. 2021.
\newblock Holistic 3d scene understanding from a single image with implicit
  representation.
\newblock In \emph{Proceedings of the {CVPR}}, pages 8833--8842.

\bibitem[{Zhao et~al.(2022)Zhao, Wei, Lin, Sun, Zhang, and Zhang}]{zhaoyu-vsd}
Yu~Zhao, Jianguo Wei, Zhichao Lin, Yueheng Sun, Meishan Zhang, and Min Zhang.
  2022.
\newblock Visual spatial description: Controlled spatial-oriented image-to-text
  generation.
\newblock In \emph{Proceedings of the {EMNLP}}.

\bibitem[{Zhou et~al.(2020)Zhou, Palangi, Zhang, Hu, Corso, and
  Gao}]{DBLP:conf/aaai/ZhouPZHCG20}
Luowei Zhou, Hamid Palangi, Lei Zhang, Houdong Hu, Jason~J. Corso, and Jianfeng
  Gao. 2020.
\newblock Unified vision-language pre-training for image captioning and {VQA}.
\newblock In \emph{Proceedings of the {AAAI}}, pages 13041--13049.

\end{thebibliography}
\bibliographystyle{acl_natbib}

\newpage

\appendix

\section{Model Specification}
Here we provide the detailed calculations of our system. For the MeanPool operation in Equation \ref{equation:g}, we calculate $\textbf{r}^G$ as:
\begin{equation}
    \begin{split}
        \textbf{r}^G = \frac{1}{M^2}\sum_{i\in G}\sum_{j\in G}\lambda_i\left(\hat{\bm{s}}^v_i \oplus \hat{\bm{s}}^e_{i,j}\right)
    \end{split}
\end{equation}
where $M$ is the node number of the \textsc{Go3D}-S$^2$G, $\lambda_i$ is a weight expansion (shown in Equation \ref{equation:g}), which could capture the subgraph features with a very high weight, $\hat{\bm{s}}^v_i$ and $\hat{\bm{s}}^e_{i,j}$ are the node and edge representations of last-layer \textsc{OcGCN}.

For VL-PTM encoding, we adopt the approach in OFA model \cite{DBLP:conf/icml/WangYMLBLMZZY22}.
The input image $I$ is first embedded by a inbuilt ResNet and flatted to a tokenized visual feature.
Then the visual features are projected to the encoder model dimension, obtaining the global visual feature $\bm{r}^I$. The process could be formalized as:
\begin{equation}
    \begin{split}
        \hat{\bm{r}}^I &= \text{ResNet}(I),\\
        \bm{r}^I &= W_v\hat{\bm{r}}^I + b_v,
    \end{split}
\end{equation}
where $W_c$ and $b_c$ are model parameters.
Then the tokenized visual features are concatenated with embedded text sequence as the inputs of VL-PTM encoder:
\begin{equation}
    \begin{split}
        \bm{h}^{I,T} &= \text{Encoder}(\bm{r}^I \oplus \bm{r}^T).
    \end{split}
\end{equation}

After that, the encoder outputs and graph feature $\bm{r}^G$ are fused through cross-attention in VL-PTM decoder. The cross-attention operation in a transformer unit could be formalized as:
\begin{equation}
    \begin{split}
        \bm{X} &= \left\{\bm{r}^G;\bm{h}^{I,T}\right\},\\
        \bm{K}=\bm{W}_k\bm{X}, \bm{V}&=\bm{W}_v\bm{X}, \bm{Q}=\bm{W}_q\bm{D_h},\\
        \textit{Attention}(\bm{Q},\bm{K}, \bm{V}) &= \textit{softmax}\left(\frac{\bm{Q}\bm{K}}{\sqrt{d}}\right)\bm{V},
    \end{split}
\end{equation}
where $\bm{r}_G$ is the graph feature, $\bm{h}^{I,T}$ is the encoder outputs, $\bm{W}_k,\bm{W}_v,\bm{W}_q$ are model parameters, $d$ is the out dimension of $\bm{W}_v$, $\bm{D}_h$ is the outputs of self-attention unit in Transformer decoder.

\subsection{3D Scene Extracting}
\label{3D Scene Extracting}

\begin{figure}[!t]
    \centering
    \includegraphics[width=0.94\linewidth]{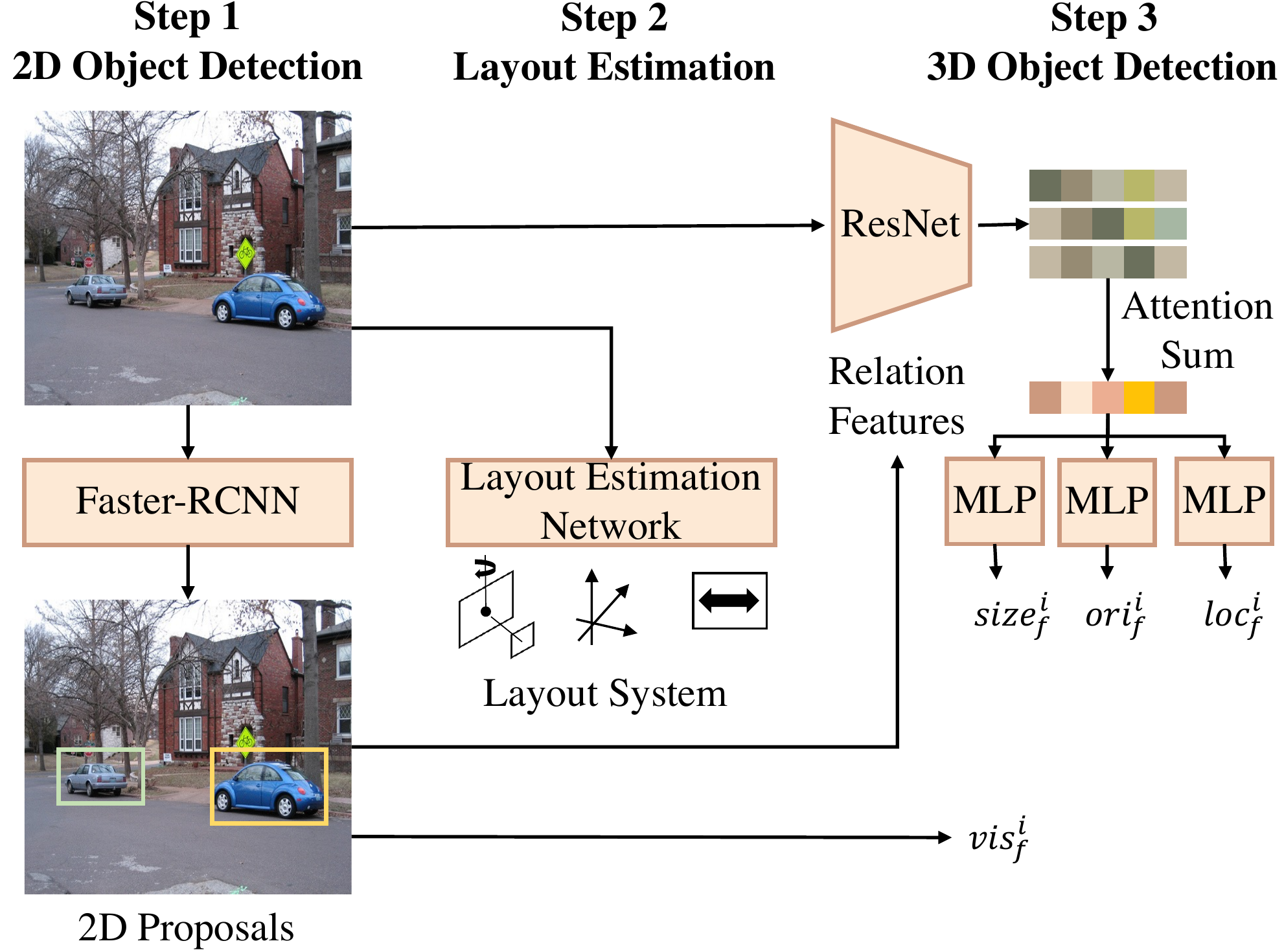}
    \caption{The framework of 3D scene extractor.}
    \label{fig:app_flowchart}
\end{figure}

\begin{figure*}[t]
    \centering
    \begin{minipage}[b]{0.32\linewidth}
        \centering
        \includegraphics[width=\textwidth]{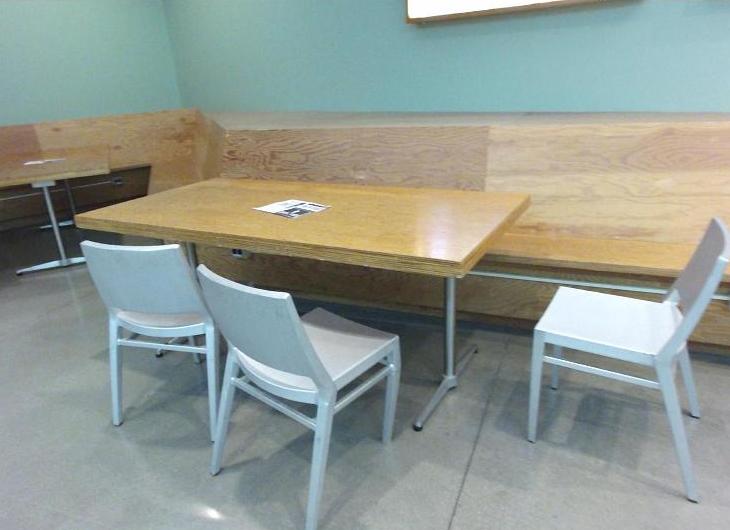}
    \end{minipage}
    \begin{minipage}[b]{0.32\linewidth}
        \centering
        \includegraphics[width=\textwidth]{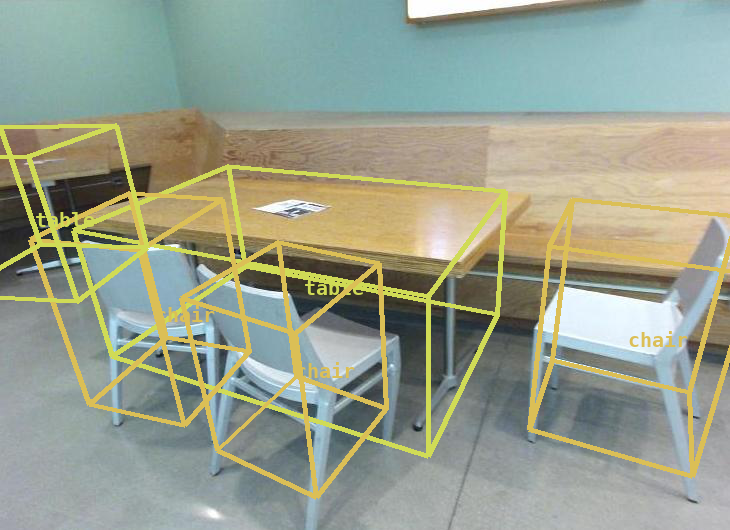}
    \end{minipage}
    \begin{minipage}[b]{0.32\linewidth}
        \centering
        \includegraphics[width=\textwidth]{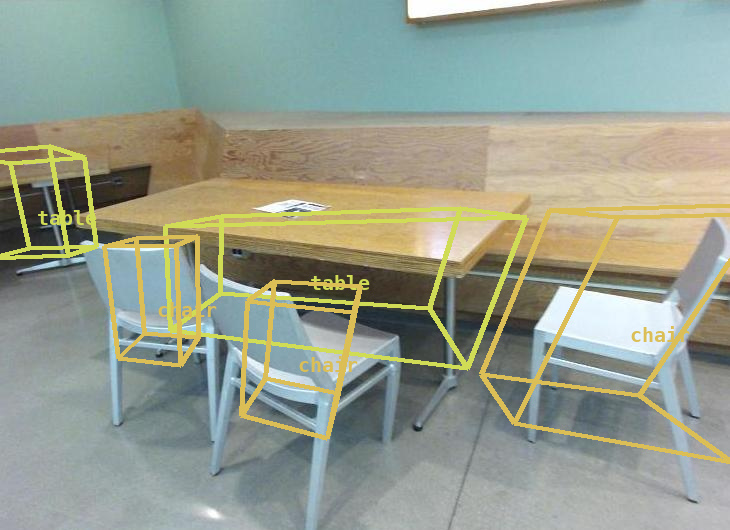}
    \end{minipage}
    \caption{Comparsion of 3D boxes between different camera intrinsics.}
    \label{fig:app_intrinsics}
\end{figure*}

We employ an external 3D scene extractor from \cite{DBLP:conf/cvpr/NieHGZCZ20}, which is pretrained on the SUN RGB-D dataset \cite{DBLP:conf/cvpr/SongLX15}. 
The model contains three modules: 1. Layout Estimation Network (LEN); 2. 3D Object Detection Network (ODN); 3. Mesh Generation Network (MGN). 
In our framework, we only use LEN and ODN, which outputs the layout coordinates system and the 3D location and pose of each object. We also need a 2D Object Detector to generate appearance feature and 2D box as the inputs for ODN. 
We directly borrow the Faster-RCNN \cite{DBLP:conf/nips/RenHGS15} to process the images to obtain the necessary RoI features, boxes, and object tags.
For the LEN, we need the camera intrinsic parameters to adjust the coordinates system. However, the VSD dataset does not provide this information of the images. We just generate a pseudo camera intrinsic matrix for all the images in our implementation. The method works in VSD because we do not need the accurate location of each object, we just need to capture the relative relations among the object pairs. Thus, the distortion of absolute coordinates will hardly impact the final results of VSD. As shown in Figure \ref{fig:app_intrinsics}, though the 3D boxes are distorted, their relative spatial relations remain unchanged.

\subsection{Spatial Scene Graph Creating}
\begin{algorithm}[t]
    \SetAlgoLined
    \KwIn{max object number $N$,\\
    \quad\quad\quad two target objects index $o_1, o_2$,\\
    \quad\quad\quad confidency of each object $f$,\\
    \quad\quad\quad centroid of each object $C$,\\
    \quad\quad\quad distance threshold $d$,\\
    \quad\quad\quad noise confidency threshold $p$}
    \KwOut{adjacency matrix $\bm{A}^{N\times N}$}
     initialization: $\bm{A}$ = \textbf{0}.
     \\
     \tcp{target object edges}
     
     $\bm{A}$[$o_1$,:] = 1, $\bm{A}$[:, $o_1$] = 1,\\
     $\bm{A}$[$o_2$,:] = 1, $\bm{A}$[:, $o_2$] = 1,\\
     \tcp{add special edges}

     \For{i in $N$}{
          \For{j in $N$}{
               dist = $||C_i - C_j||$\\
               \If{dist > $d$}{
                    $\bm{A}_{ij}$ = 1
               }
          }
     }

     \tcp{remove noise objects}

     \For{i in $N$}{
          \If{$f_i < p$ and $o_i$ is not target object}{
               $\bm{A}$[i,:] = 0, $\bm{A}$[:,i] = 0
          }
     }
     
     \caption{G\textsc{o}3D-S$^2$G Creating}
     \label{alg:1}
\end{algorithm}
The detailed algorithm of S$^3$ is shown in Algorithm \ref{alg:1}. Suppose that the max object number is $N$ ($N$=36 in our implementation). We first initialize an $N\times N$ adjacency matrix $\bm{A}=\bm{0}$. When adding the Target-pair and Target-surrounding edges, we set the columns and rows of the target object index to 1. When adding the Near-neighbor edge, we traverse the edges and set the one with $dist$ > $d$ in $\bm{A}$ to 1. At last, we should remove the edges of noisy objects, setting the related items of \textbf{A} to 0.

\subsection{Overall Training Procedure}
We have several pretrained external modules in our system. 
First, we prepare the pretrained ResNet for image embedding, which will be used in 3D layout estimation, 3D object detection and the OFA image embedding. Before training, we use the pretrained Faster-RCNN to preprocess all the images for 2D object detection, obtaining the 2D boxes and RoI features. 
We also prepare the 3D scene extractor pretrained on SUNRGB-D with the method in \cite{DBLP:conf/cvpr/NieHGZCZ20}.

Then we collect the three modules and train our whole system, where the 3D extractor and the decoder (VL-PTM) are initialized by pretrained parameters.
The \textsc{OcGCN} and other parameters (e.g. edge embedding, connecting-strength scorer) are initialized randomly. 
Before global training, we pretrain the VL-PTM with direction terms as Table \ref{tab:direction_mapping}.
During global training, the 3D scene extractor are frozen while others will be updated.

We have two main training targets, e.g., the cross-entropy between Vl-PTM decoder outputs and the ground-truth description of VSD $\mathcal{L}_t$, and the cross-entropy between connecting-strength score distribution and target object index $\mathcal{L}_c$:
\begin{equation}
    \begin{split}
        \mathcal{L}_t &= -\sum_{l=1}^{N_w}\log p_\theta\left(y_l|\bm{x},\bm{y}_{<l}\right),\\
        \mathcal{L}_c &= -\sum_{i=1}^{N}\delta(i\in O_g)(\log a_{t_1,i}+log a_{t_2,i}),
    \end{split}
\end{equation}
where $N_w$ is the text length, $\bm{x}$ is the VL-PTM encoder outputs and $y_l$ is decoder outputs in the $l$-th step ,
$N$ is the node number of \textsc{Go3D}-S$^2$G, $a_{i,j}$ is connecting-strength score during S$^3$,
$t_1,t_2$ are the two target objects,
$O_g$ is a set that contains the objects appear in the ground-truth description, and $\delta\left(\cdot\right)$ is the two value function that outputs 1 when $\cdot$ is true, otherwise 0.
We obtain $O_g$ by string searching in the descriptions.
Then the final loss is $\mathcal{L} = \mathcal{L}_t + \mathcal{L}_c$.

\section{Experiment Specification}

\subsection{Data Analyses}
\pgfplotsset{compat=1.7,every axis title/.append style={at={(0.5,-0.45)}, font=\fontsize{14}{1}\selectfont},every axis/.append style={xtick pos=left,ytick pos=left,tickwidth=1.5pt}}
\usetikzlibrary{matrix}
\usepgfplotslibrary{groupplots}
\usetikzlibrary{patterns,backgrounds}

\definecolor{c1}{RGB}{252,232,212}
\definecolor{c2}{RGB}{184,183,163}
\definecolor{c3}{RGB}{107,112,092}
\definecolor{c4}{RGB}{203,153,126}
\definecolor{c5}{RGB}{107,112,092}

\begin{figure*}[t]
\centering
\begin{tikzpicture}
\begin{axis}[
	ybar,
	ytick = {400, 1000, 1600, 2200},
	yticklabels={400, 1000, 1600, 2200},
	ymax=2200,ymin=400,
	y tick label style = {yshift=-0.5em, text height=0ex,font=\scriptsize},
    x label style = {font=\scriptsize},
    y label style = {yshift=-0.5em,font=\scriptsize},
	axis line style={-},
    axis x line*=bottom,
    axis y line*=left,
	enlargelimits=0.05,
	legend style={at={(0.02,0.99)},anchor=north west, draw=none, legend columns=-1, font=\scriptsize},
	xticklabels={man, building, tree, table, window, wall, woman, car, water, sky, fence, person, plant, grass, chair, sign, train, road, plate, street},
	xtick={1,2,3,4,5,6,7,8,9,10,11,12,13,14,15,16,17,18,19,20},
	xmax=20, xmin=1,
	x tick label style = {yshift=0.05em,rotate=90, align=center,font=\small},
	title style={yshift=-0.9em,font=\small},
    ymajorgrids,
	width = \textwidth,
	height = 5cm,
	]
	\addplot[c4, pattern= north east lines, thick, pattern color=c4, bar shift=0pt, bar width = 1em] coordinates
	{
		(1,2162)
        (2,1840)
        (3,1378)
        (4,1108)
        (5,1103)
        (6,964)
        (7,935)
        (8,839)
        (9,774)
        (10,758)
        (11,638)
        (12,626)
        (13,622)
        (14,617)
        (15,615)
        (16,603)
        (17,562)
        (18,538)
        (19,478)
        (20,469)
	};

	
\end{axis}

	


\end{tikzpicture}
\caption{Top 20 objects distribution in VSD dataset.}
\label{fig:tag_dist}
\end{figure*}
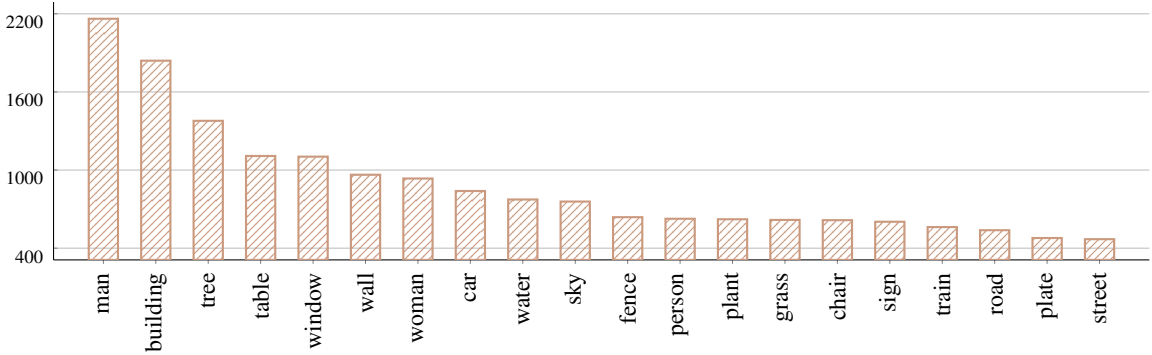
\begin{table}[t]
    \fontsize{10}{12.5}\selectfont
    \setlength{\tabcolsep}{1.mm}
    \begin{center}
    \begin{tabular}{l}
    \hline
     \bf Object Types \\
    \hline
    wall, floor, cabinet, bed, chair, \\
    sofa, table, door, window, bookshelf, \\
    picture, counter, blinds, desk, shelves, \\
    curtain, dresser, pillow, mirror, floor, \\
    clothes, ceiling, books, refridgerator, television, \\
    paper, towel, shower curtain, box, whiteboard, \\
    person, night stand, toilet, sink, lamp, \\
    bathtub, bag, other\\
    
    \hline
    \end{tabular}
    \caption{Objects in SUNRGB-D dataset.}
    \label{tab:3dobj}
    \end{center}
    \end{table}
The VSD dataset contains images from Visual Genome, Flickr, and NYU-Depth \cite{zhaoyu-vsd,DBLP:conf/iccv/YangRD19}, where NYU-Depth is the indoor dataset \cite{DBLP:conf/eccv/SilbermanHKF12}.

Figure \ref{fig:tag_dist} shows the distribution of object tags in the VSD dataset. Table \ref{tab:3dobj} lists the object types in SUNRGB-D dataset which are used to pretrain 3D scene extractor. There are several common types, such as ``person'', ``table'', ``chair'', ``wall'' and ``window''. Thus the 3D scene extractor could achieve a comparable performance in VSD. For other object types, the built-in 2D object detector (pretrained on VG or COCO) could provide a passable 2D location information.

\begin{figure}[t]
    \centering
    \includegraphics[width=0.98\linewidth]{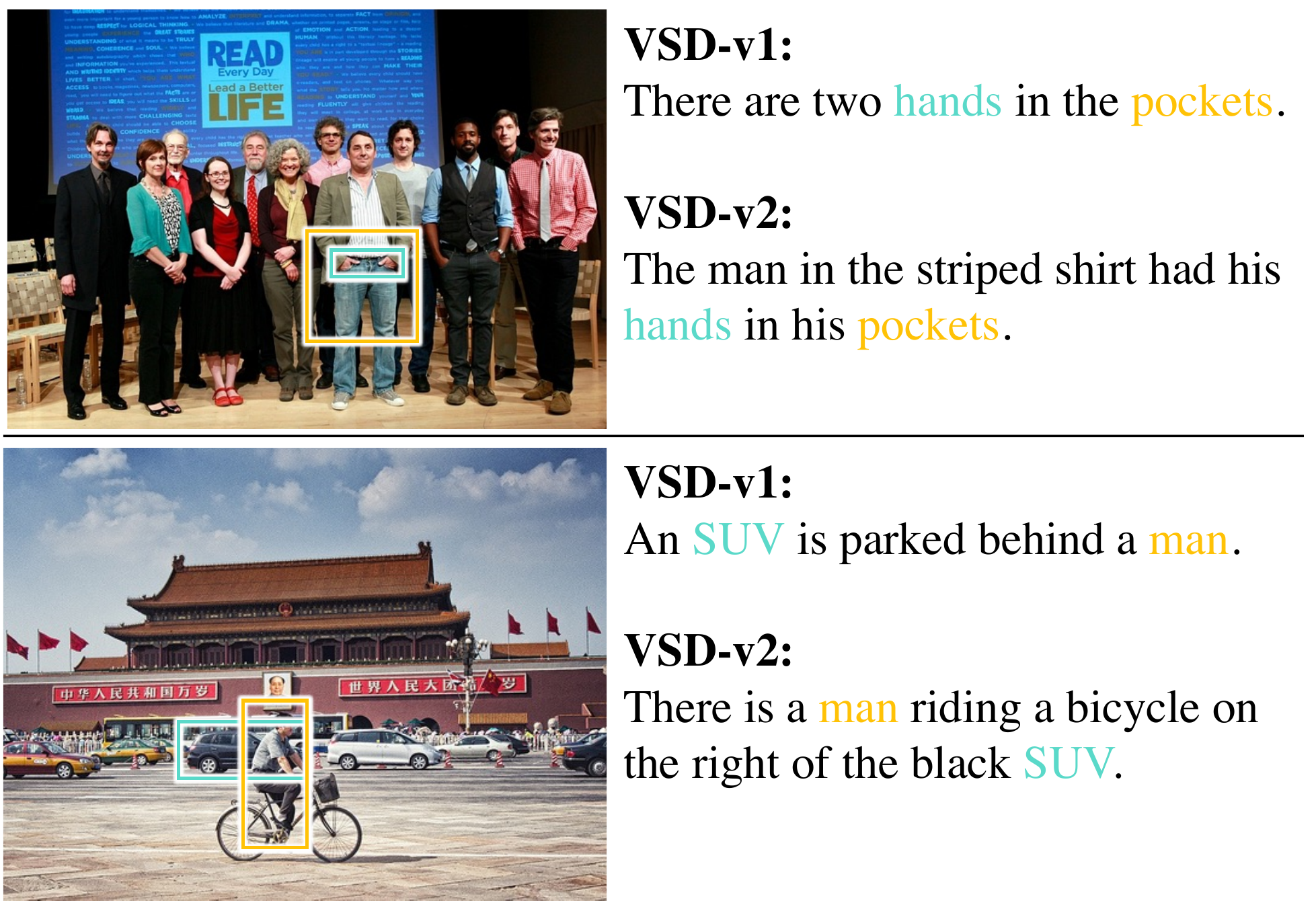}
    \caption{Comparison between annotations of VSD-v1 and VSD-v2.}
    \label{fig:app_data_compare}
\end{figure}

There are two versions of VSD dataset released, where the VSD-v1 contains large-scale while relative simple descriptions and the VSD-v2 contains small-scale while diversified descriptions. 
The two versions share the same image set.
The statistics of the two datasets are shown in Table \ref{tab:app_data}. Figure \ref{fig:app_data_compare} shows the difference between the annotations of the two datasets. The annotated sentence in VSD-v1 is relative short and simple while those in VSD-v2 contains more semantics and be more challenging.

\subsection{Extended Experimental Implementations}
\paragraph{Data Preprocess} For memory space and time saving, we preprocess the data through Faster-RCNN to obtain the RoI featues, object tags and 2D boexs and save them to files. For 3D scene extractor, we also preprocess the images, saving their ResNet features to files.

\setlength{\tabcolsep}{2pt}
\begin{table}[t]
\small
\begin{center}
\begin{tabular}{cccccc}
\hline
\multirow{2}{*}{} & \multirow{2}{*}{\#Img} & \multicolumn{2}{c}{VSD-v1} & \multicolumn{2}{c}{VSD-v2}\\
\cmidrule(r){3-4} \cmidrule(r){5-6}
& & \#SENT & AvgLEN & \#SENT & AvgLEN\\
\hline
Train    & 20,490   & 116,791 & 7.35 & 22101 & 8.88\\
Dev    & 2,927  & 16,823  & 7.33 & 3140 & 8.85\\
Test & 5,855 & 10,038 & 8.04 & 5.34 & 9.00 \\
\hline
\end{tabular}
\caption{The statistics of the two versions of VSD dataset.}
\label{tab:app_data}
\end{center}
\end{table}

\paragraph{VL-PTM Configuration} For fairly comparison, we employ the OFA-base as our VL-PTM, which has the similar scale of parameters to the baselines. The layers number is 6 for both encoder and decoder, and the head number of the multi-head attention is 6 as well.
Besides OFA, we can use any type of VL-PTMs such as VLT5/VLBart, or combination of single VL encoder and text decoder, such as ``CLIP-GPT'', ``ViT-GPT''. For more comparison, we leave them in future works.

\begin{table*}[!t]
    \fontsize{10}{12.5}\selectfont
    \setlength{\tabcolsep}{1.mm}
    \begin{center}
    \begin{tabular}{clcccccccccc}
    \hline
    \multicolumn{2}{c}{} & \multicolumn{3}{c}{K=5 Samlpling} & \multicolumn{3}{c}{K=10 Samlpling} \\
    \multicolumn{2}{c}{} & mBLEU-4$\downarrow$ & BLEU-4@K$\uparrow$ & SPICE@K$\uparrow$ & mBLEU-4$\downarrow$ & BLEU-4@K$\uparrow$ & SPICE@K$\uparrow$ \\
    \hline
    \multicolumn{8}{l}{\textbf{Beam Search}} \\
    & VL-T5 & 8.62 & 33.02 & 60.55 & 8.02 & 33.27 & 61.11 \\
    & OFA & 7.77 & 32.72 & 60.37 & 7.41& 33.09 & 61.06 \\
    & Ours & 7.6 & 33.12 & 60.67 & 7.38 & 33.46 & 61.31 \\
    \hline
    \multicolumn{8}{l}{\textbf{Scene Subgraph Sampling}} \\ 
    & Ours & 5.01 & 34.44 & 61.99 & 4.83 & 34.75 & 62.61 \\
    \hline
    \end{tabular}
    \caption{Diversity evaluation results in VSD-v2.
    }
    \label{tab:diversity_app}
    \end{center}
\end{table*}

\subsection{Hyperparameters}
\label{app:hyperparameter}
\begin{table}[t]
    \fontsize{10}{12.5}\selectfont
    \setlength{\tabcolsep}{1.mm}
    \begin{center}
    \begin{tabular}{lc}
    \hline
     \bf Hyper-param. & \bf Value \\
    \hline
    dimension of ROI feature & 2048 \\
    dimension of OFA hiddens & 768 \\
    dimention of edge feature & 64 \\
    layer number of GCN & 3 \\
    max object number & 36 \\
    distance threshold $d$ & 0.2 \\
    confidency threshold $p$ & 0.7 \\
    S$3$ cutting threshold $p_{cut}$ & 0.1 \\
    max text length & 40 \\
    optimizer & AdamW \\
    warm up ratio & 0.05 \\
    weight decay & 0.01 \\
    Adam $\epsilon$ & 1e-6 \\
    Adam $\beta_1$ & 0.9 \\
    Adam $\beta_2$ & 0.999 \\
    dropout & 0.1 \\
    learning rate & 5e-5 \\
    bach size & 16 \\
    epoch & 30 \\
    beam search number & 5 \\

    \hline
    \end{tabular}
    \caption{Model hyperparameters.}
    \label{tab:hyperparameter}
    \end{center}
    \end{table}
Table \ref{tab:hyperparameter} lists the hyperparameters of our implementation.

\subsection{Baselines}
We use some strong image-to-text models as our baselines.
\begin{compactitem}
    \item \textbf{Oscar} \cite{DBLP:conf/eccv/Li0LZHZWH0WCG20} is a BERT-like visual-language pretrained model, only containing the Transformer encoder. Oscar takes tokenized RoI features of the image as visual inputs. This model has shown strong performance on Image Captioning and VQA.
    \item \textbf{VLT5/VLBart} \cite{DBLP:conf/icml/ChoLTB21} are continue pretrained on image-to-text tasks from generative encoder-decoder language model Bart and T5. Previous work \cite{zhaoyu-vsd} has proven their capability on solving VSD task.
    \item \textbf{OFA} \cite{DBLP:conf/icml/WangYMLBLMZZY22} is a new vision-language model, which are pretrained by a mount of vision-language multi-modal tasks. It has achieved several SoTA results on existing image-to-text tasks.
\end{compactitem}

\section{Extended Experiments}

\subsection{Diversity Evaluation}
\begin{figure*}[t]
    \centering
    \includegraphics[width=0.98\linewidth]{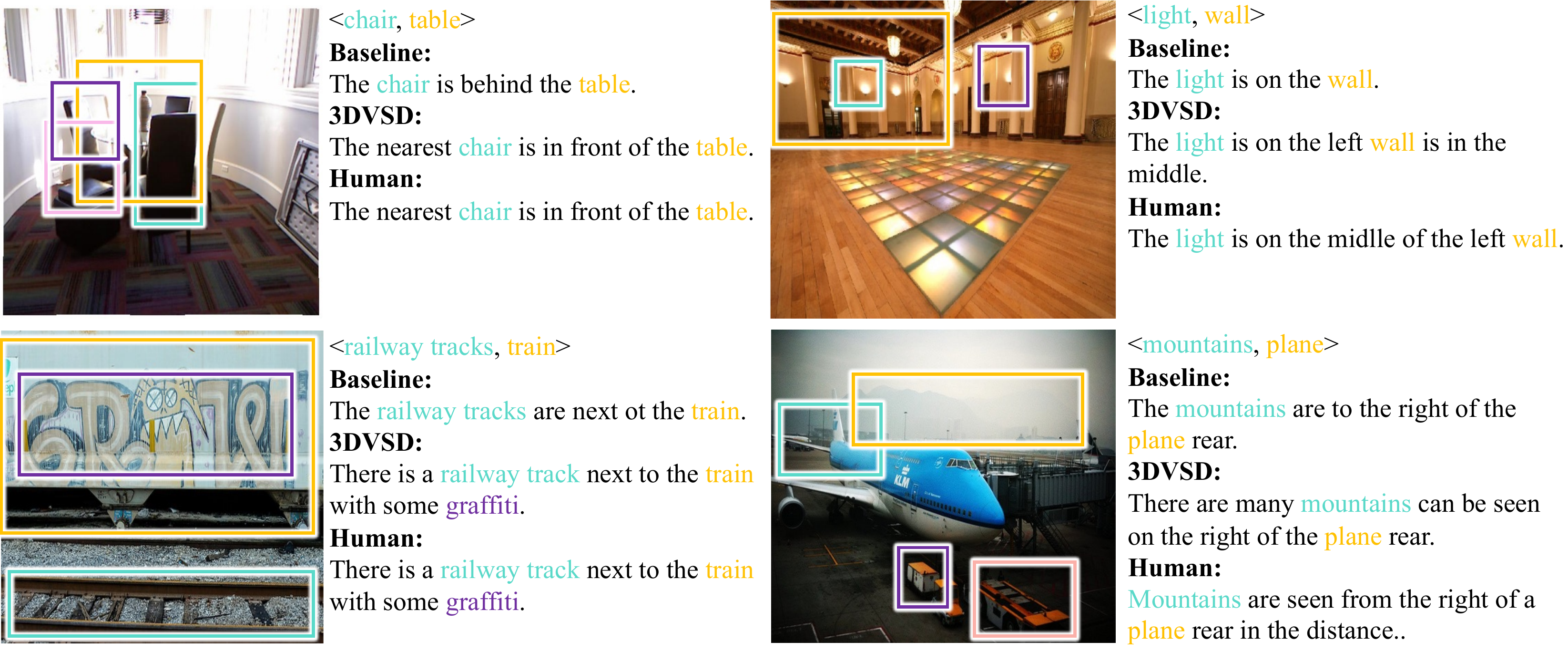}
    \caption{Cases comparison among models and human.}
    \label{fig:app_case}
\end{figure*}
We compare more results of sampling scale for diversity evaluations. Figure \ref{tab:diversity_app} reports the results of K=5 and K=10 on VSD-v2. The tendency of K=10 is consistent with that of K=5, where the S$^3$ achieves lower mBLEU-4 and higher BLEU-4@K and SPICE@K.

For human evaluation, our 5-point Likert scale is designed as follows:
\begin{compactitem}
    \item \textbf{Spatial Accuracy}: The sentences correctly describe the spatial relationship of the target objects.
    \item \textbf{Spatial Diversity}: These sentences describe diversified spatial semantics.
    \item \textbf{Fluency}: The sentences are readable and not different from human sentence-making.
\end{compactitem}
Each question will be answered by a number from 1 to 5, denoting ``Strongly Disagree'', ``Disagree'', ``Neither disagree nor agraa'', ``Agree'' and ``Strongly agree''.
We select 100 samples and generated 5 sentences for each, and sent them to the evaluators for scoring.
\subsection{More Case Study}
Figure \ref{fig:app_case} provides some more qualitative comparisons among models and human.
Compared with baselines, our model could generate description more like human-made.

\end{CJK*}
\end{document}